\documentclass{article}
\newcommand{\CIMArxivVersion}{}
\AddToHook{begindocument/before}{%
\iclrfinalcopy
\author{%
\begin{minipage}[t]{\dimexpr\textwidth-2\tabcolsep\relax}
\centering\normalfont
Yinghua Xu\textsuperscript{1,2}\quad
Pengyang Xie\textsuperscript{2}\thanks{Corresponding author}\quad
Yanxiang Zhou\textsuperscript{2}\quad
Chuanzhi Tong\textsuperscript{2}\\[3pt]
Xiao Xiao\textsuperscript{2}\quad
Chaoqun Feng\textsuperscript{2}\quad
Xiaoqing Wang\textsuperscript{2}\\[7pt]
{\small\textsuperscript{1}College of Control Science and Engineering, Zhejiang University}\\[2pt]
{\small\textsuperscript{2}Alibaba International Digital Commerce}\\[7pt]
{\footnotesize\texttt{22432113@zju.edu.cn}}\\[2pt]
{\footnotesize\texttt{\{xiepengyang.xpy,zhouyanxiang.zyx,tongchuanzhi.tcz\}@alibaba-inc.com}}\\[2pt]
{\footnotesize\texttt{\{wanbao.xiaoxiao,fengchaoqun.fcq,robin.wxq\}@alibaba-inc.com}}
\end{minipage}%
}
\hypersetup{%
  hidelinks,
  pdftitle={CIM-Embed: Learning Instruction-Conditioned Multimodal Embeddings for Clustering},
  pdfauthor={Yinghua Xu, Pengyang Xie, Yanxiang Zhou, Chuanzhi Tong, Xiao Xiao, Chaoqun Feng, Xiaoqing Wang}%
}
}
\AddToHook{cmd/maketitle/after}{\lhead{Preprint}}
\usepackage[T1]{fontenc}
\usepackage{iclr2027_conference,times}

\usepackage{amsmath,amsfonts,bm}

\def\eqref#1{equation~\ref{#1}}

\def\1{\bm{1}}

\DeclareMathAlphabet{\mathsfit}{\encodingdefault}{\sfdefault}{m}{sl}
\SetMathAlphabet{\mathsfit}{bold}{\encodingdefault}{\sfdefault}{bx}{n}

\makeatletter
\renewcommand{\eqref}[1]{\textup{\tagform@{\ref{#1}}}}
\makeatother
\usepackage{amssymb,placeins,float}
\usepackage{hyperref}
\usepackage{url}
\usepackage{booktabs}
\usepackage[normalem]{ulem}
\usepackage{tabularx}

\usepackage{graphicx}
\usepackage{xcolor}
\title{CIM-Embed: Learning\\Instruction-Conditioned Multimodal\\Embeddings for Clustering}
\begin{document}
\maketitle
\begin{abstract}
Multimodal objects can require different partitions under different instructions, as when products in an e-commerce catalog are grouped by color, material, or both. Retrieval-oriented embeddings may not recover these partitions, while generative clustering often relies on intermediate descriptions or iterative decisions. This paper presents \textbf{CIM-Embed}, a framework for learning instruction-conditioned multimodal embeddings for clustering. Supervised contrastive learning uses same-group positives and negatives from other groups, optionally including other tasks; heterogeneous structured batches accommodate tasks with different group counts. At inference, the model maps each object's image, text, and instruction directly to an embedding; HDBSCAN then forms a partition without the ground-truth cluster count. We construct \textbf{ProductMIC} from product metadata in Amazon Reviews 2023, comprising training data and an evaluation benchmark for instruction-conditioned clustering. The benchmark contains 5,373 products and 449 single-attribute and compositional grouping tasks spanning 50 leaf categories excluded from our training. With full-dimensional embeddings and HDBSCAN, direct within-task SupCon training achieves a mean per-task V-measure of 95.01, compared with 53.06 for WeMM 9B, the strongest general-purpose embedding baseline evaluated. Frozen product-trained encoders also improve out-of-domain conditional grouping, while controlled transfer experiments distinguish partition transfer from instruction response. We will release the code and dataset upon acceptance.
\end{abstract}

\section{Introduction}
\label{sec:introduction}

Organizing products in e-commerce often requires grouping the same collection by different criteria, such as color, material, or their combination. This need reflects a broader problem: for the same collection of multimodal objects, the appropriate partition depends on the current grouping criterion, and whether two objects belong to the same group can change accordingly. For example, red and blue wooden cups belong together when grouped by material but separate when grouped by color; grouping by both color and material requires agreement on both attributes. Evidence for the relevant attributes may be distributed across images and text. We address \textbf{instruction-conditioned multimodal clustering}, investigating how to learn multimodal representations that enable a clustering algorithm to recover the partition specified by a natural-language grouping instruction.

\begin{figure}[t]
\centering
\includegraphics[width=\linewidth,height=0.48\textheight,keepaspectratio]{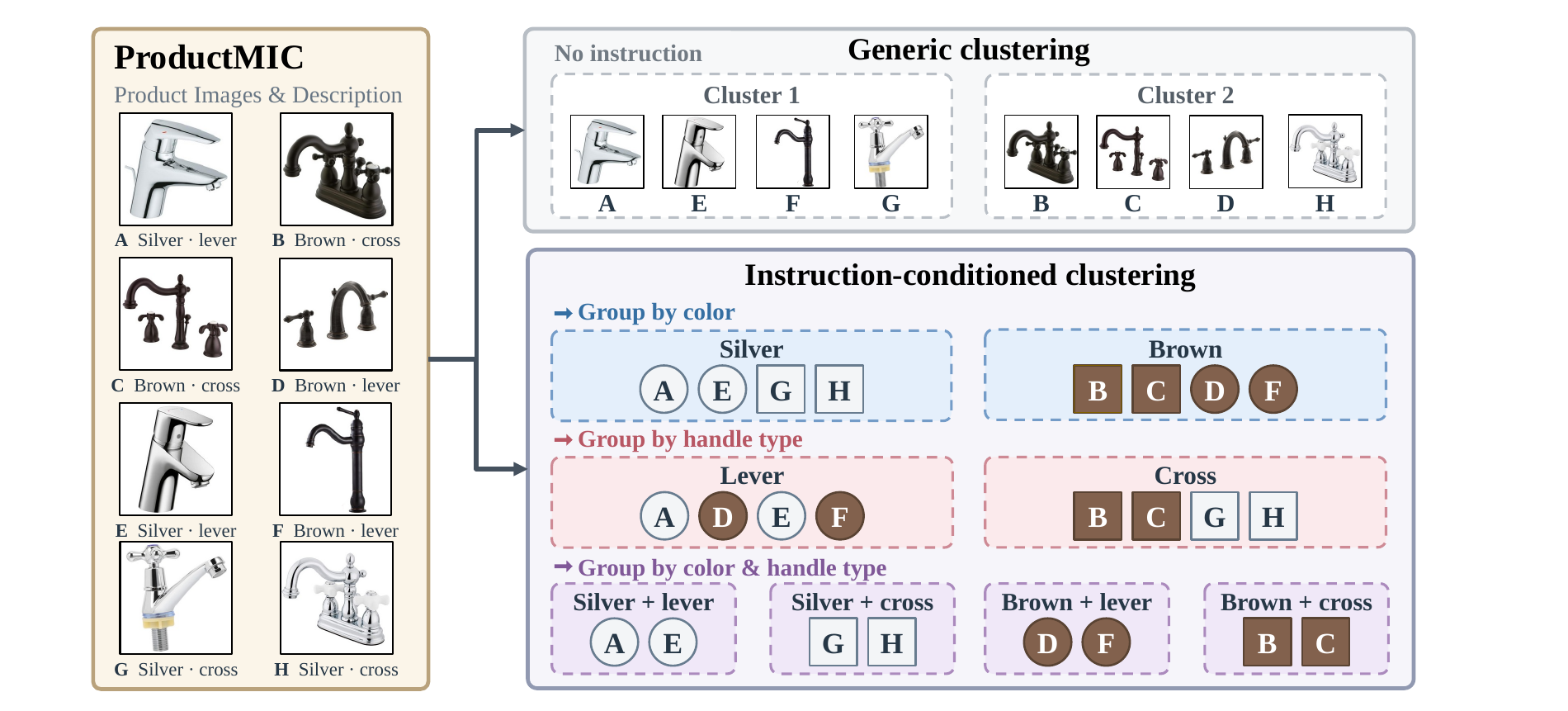}
\caption{Illustration of grouping the same eight faucets by color, handle type, or both. All partitions are schematic rather than empirical model predictions.}
\label{fig:instruction-overview}
\end{figure}

Shared multimodal embedding spaces support image--text alignment and retrieval, while instruction-aware models extend these capabilities to task-conditioned embedding tasks~\citep{radford2021clip,uniir2024,mmembed2025,qwen3vlembedding2026,zhou2026wemmembedding}. Retrieval evaluates whether relevant objects rank ahead of alternatives for an anchor, while conditional similarity evaluates whether similarities reflect a specified criterion. Neither evaluation directly establishes whether a clustering algorithm can recover the complete instruction-specified partition of an object collection. In the task studied here, the same pair of objects may belong together under one grouping instruction and to different groups under another; success is assessed by recovering the target partition across the collection. Conditional clustering systems offer another route: the framework built around GCPM generates concept descriptions, clusters their embeddings, and uses language-model-guided cluster merging~\citep{zhong2026guidelineclustering}. This motivates a direct representation-learning question: can the embedding geometry itself be trained to make the partition requested by the instruction recoverable?

We introduce \textbf{CIM-Embed} (Clustering-oriented Instruction-conditioned Multimodal Embeddings), which adapts Qwen3-VL-Embedding using instruction-defined group relations~\citep{qwen3vlembedding2026}. Within a task, objects assigned to the same target group provide multiple positives for supervised contrastive learning, while objects in different groups provide negatives~\citep{khosla2020supervised}. To supply these relations at each training step, partition-structured batches contain multiple groups and multiple objects per group and can combine tasks with different group counts. We study two comparison scopes: Within compares instances only inside the same task, whereas Cross additionally uses instances from other tasks in the physical batch as negatives. Gradient analysis characterizes how these scopes affect the contrastive objective. At inference, the model directly embeds each object together with its instruction, and HDBSCAN recovers a partition without a supplied cluster count~\citep{mcinnes2017hdbscan}; no intermediate descriptions or language-model-based merge decisions are required.

We instantiate this task in e-commerce by constructing \textbf{ProductMIC} from Amazon Reviews 2023~\citep{hou2026bridging}. It provides training supervision for grouping by individual attributes and by combinations of attributes, together with a benchmark of 449 tasks over 5,373 products from 50 leaf categories excluded from our training. With full-dimensional embeddings and HDBSCAN, direct within-task SupCon training achieves a mean per-task V-measure of 95.01, versus 53.06 for WeMM 9B, the strongest general-purpose embedding baseline evaluated (Table~\ref{tab:embedding-main}). As an extension for category retention, staged training adds generic adaptation and replay; its Cross SupCon configuration achieves 93.85 on conditional grouping and 97.08 on category clustering (Tables~\ref{tab:embedding-main} and~\ref{tab:embedding-generic}). We also evaluate frozen encoders on Cards, CLEVR-4, and CIFAR10-MC to examine cross-domain grouping and instruction response (Section~\ref{sec:rq5}; Appendix~\ref{app:additional-transfer}).

Our contributions are:
\begin{itemize}
    \item We present CIM-Embed, a framework that learns image--text embeddings for direct clustering under natural-language grouping instructions. Instruction-defined group membership supplies multi-positive supervision, while structured batches enable joint training across tasks with different numbers of groups.
    \item To train and evaluate instruction-defined partition recovery, we construct ProductMIC, comprising training data and a benchmark of 449 single-attribute and compositional grouping tasks across 50 leaf categories excluded from training. The benchmark supports comparisons with general-purpose multimodal embeddings, conditional representation methods, and generative clustering systems.
    \item We examine within-task partition recovery, the organization of conditioned tasks in a shared space, and category-structure retention through comparison-scope and staged-training experiments. Exact gradient decomposition characterizes the objective differences between Within and Cross. Cross-domain matched controls further show that improved partition quality does not necessarily imply reliable instruction switching.
\end{itemize}

\section{Related Work}
\label{sec:related-work}

\paragraph{Multimodal Embeddings and Instruction-Aware Representations.}

Multimodal embedding methods learn which information to preserve for a task. UniIR and MM-Embed focus on shared representations for cross-modal alignment and retrieval~\citep{uniir2024,mmembed2025}. VLM2Vec and Qwen3-VL-Embedding incorporate task instructions, adapting representations to retrieval, classification, and other embedding tasks~\citep{jiang2025vlm2vec,qwen3vlembedding2026}.

These models use instructions to emphasize task-relevant semantics, typically under similarity or retrieval objectives. CIM-Embed instead trains task-conditioned multimodal embeddings for recovery of instruction-specified groups.

\paragraph{Conditional Similarity and Controllable Representations.}

Conditional similarity methods make proximity depend on an external criterion. Conditional subspaces select representations by semantic condition~\citep{veit2017csn}; C-STS and GeneCIS study conditioned textual similarity and visual retrieval~\citep{deshpande2023csts,vaze2023genecis}. ABC adapts visual models to instructions, while CLAY transforms existing visual embeddings~\citep{schneider2025abc,lim2026clay}. InBedder and GSTransform study instruction-aware text embeddings, including downstream clustering~\citep{peng2024inbedder,feng2025gstransform}.

The shared goal is condition-dependent proximity. CIM-Embed evaluates this control through \textbf{partition recovery}: jointly encoding images and text to organize an object collection under single-attribute or compositional instructions. This tests whether conditioned similarities support a complete target partition.

\paragraph{Conditional and Generative Clustering.}

In clustering, semantic control does not have to be implemented through the embedding model itself. Many methods instead introduce additional semantic representations or reasoning steps into the clustering pipeline. Multi-MaP and Multi-Sub construct proxy representations for different semantic perspectives~\citep{yao2024multimap,yao2024multisub}; IC$|$TC and ITGC use generated descriptions or concepts to guide grouping~\citep{kwon2024ictc,zhao2025itgc}; other approaches inject semantic constraints through text rewriting, relational feedback, or generated cluster assignments~\citep{pangnaylor2025controllable,zhang2023clusterllm,qing2026clusterr1}. GCPM further combines guideline-aware concept descriptions, embedding-based clustering, and language-model-guided cluster merging~\citep{zhong2026guidelineclustering}.

These methods demonstrate that generation and semantic reasoning can provide effective control over clustering, but the conditioning signal mainly acts through the clustering pipeline or intermediate semantic representations. CIM-Embed takes a different route: the grouping condition is incorporated into the object representation before clustering. At inference time, each image--text object is encoded together with the instruction, after which a standard clustering algorithm produces the partition, without requiring intermediate description generation or language-model-based cluster refinement.

\paragraph{Contrastive Supervision and Batch Construction.}

In supervised contrastive and metric learning, positive selection and within-batch competition shape representation geometry. InfoNCE typically uses a single positive per anchor~\citep{oord2018cpc}, whereas supervised contrastive learning supports multiple same-class positives~\citep{khosla2020supervised}. Class-balanced sampling, within-batch mining, and pair selection determine the supervision relations available during optimization~\citep{hermans2017defense,wang2019multisimilarity}.

Our contribution lies in how contrastive supervision is instantiated for instruction-dependent partition learning: the supervision relation itself is \textbf{defined by the current instruction}. The same pair of objects may form a positive pair under one grouping instruction but belong to different groups under another. To expose these dynamic relations reliably, we construct partition-structured batches containing multiple target groups and multiple objects per group, while allowing heterogeneous grouping tasks to share the same physical batch. Within and Cross then control whether contrastive competition remains inside a task unit or extends across task units, and we analyze how this change in comparison scope affects the resulting gradients and conditional embedding geometry.

\section{Instruction-Conditioned Multimodal Clustering}
\label{sec:benchmark}
\label{sec:task-definition}

\paragraph{Task definition}
Consider a pool $\mathcal X=\{x_i\}_{i=1}^{n}$ of multimodal objects within a task-defined scope $r$; each $x_i=(v_i,t_i)$ contains an image and text. For the attribute-based tasks studied here, a natural-language instruction $c$ selects the attributes $S$ used for grouping within this pool, such as $S=\{\text{color},\text{material}\}$. We index the task by $U=(r,S)$: $r$ specifies which objects are eligible, whereas $S$ specifies how they should be grouped. For each selected attribute $k$, let $\ell_{ik}$ denote object $i$'s normalized attribute label. The tuple $y_i^U=(\ell_{ik})_{k\in S}$ defines the target co-membership relation within the task pool:
\begin{equation}
i\sim_U j\quad\Longleftrightarrow\quad y_i^U=y_j^U,\qquad x_i,x_j\in\mathcal X.
\label{eq:target-relation}
\end{equation}
This equivalence relation defines the target partition $\Pi_c$, grouping objects whose values agree on every instructed attribute. Holding the object pool fixed while changing $S$ can yield different partitions.

\paragraph{Conditional encoding and clustering}
The multimodal embedding model processes each object independently as $\bm z_i^c=f_\theta(v_i,t_i,c)\in\mathbb R^d$. Let $\operatorname{Clust}_\lambda$ denote a clustering algorithm with hyperparameters $\lambda$, producing $\widehat\Pi_c=\operatorname{Clust}_\lambda(\{\bm z_i^c\}_{i=1}^n)$. We use HDBSCAN without providing the ground-truth cluster count and allow noise assignments~\citep{mcinnes2017hdbscan}. The goal is to recover the target partition specified by the instruction. Each embedding depends only on the corresponding object and instruction, whereas HDBSCAN uses all embeddings in the task pool jointly.

\paragraph{ProductMIC: Instruction-Conditioned Clustering Benchmark}
We construct ProductMIC to support this task by pairing product images and text with natural-language grouping instructions. Each ProductMIC task groups products of one type according to the attributes specified by its instruction. We normalize attribute values and assign products with identical values for all instructed attributes to the same group. These group labels provide contrastive supervision in ProductMIC-Train and reference partitions in ProductMIC-Bench. Every product included in a task has an accepted label for each attribute specified by that task's instruction.

ProductMIC-Bench contains 449 single-attribute and compositional grouping tasks over 5,373 distinct products, totaling 14,997 product--task instances. All 50 evaluation leaf categories are excluded from our training sources. The model receives product images, titles, descriptive text, and grouping instructions, without target attribute values or group labels. Appendix~\ref{app:data-construction} details task composition, pool sizes, input fields, and split rules; Appendix~\ref{sec:matched-pools} describes instruction comparisons on fixed object pools.

\begin{figure}[!t]
\centering
\includegraphics[width=\linewidth,height=0.48\textheight,keepaspectratio]{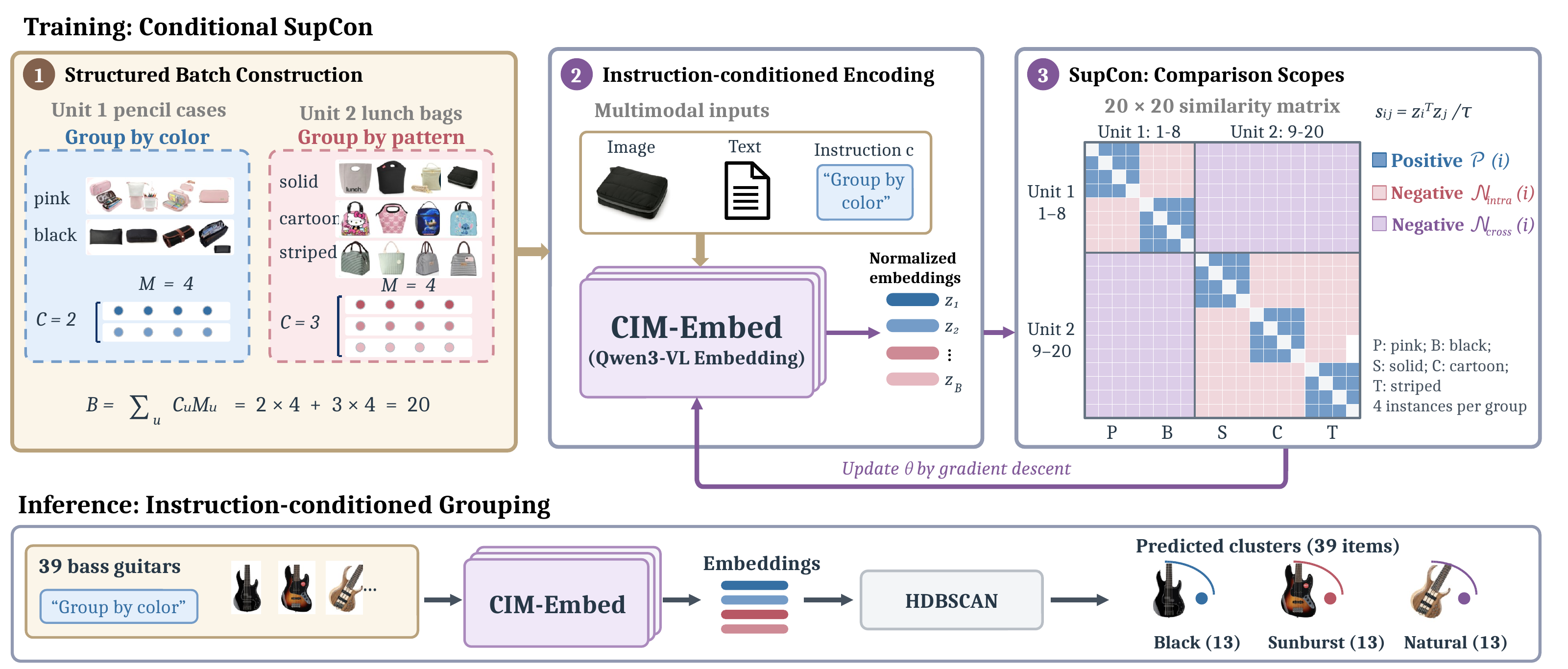}
\caption{CIM-Embed training and inference. The illustrated batch combines two tasks with 20 object--instruction instances. Within SupCon compares groups within each task; Cross SupCon additionally includes instances from other tasks. At inference, image, text, and instruction produce embeddings for HDBSCAN. Partitions are schematic.}
\label{fig:training-inference}
\end{figure}

\section{Learning Instruction-Conditioned Multimodal Embeddings}
\label{sec:method}

CIM-Embed uses the target co-membership relation defined by the grouping instruction to construct contrastive supervision (Figure~\ref{fig:training-inference}). The same pair of objects may be positive under one instruction and negative under another. We instantiate these task-dependent relations with standard supervised contrastive learning. Partition-structured sampling supplies multiple target groups and multiple objects per group within each task unit, exposing both same-group support and competing groups at each update. When several task units share a physical batch, Within and Cross determine whether contrastive competition is restricted to the current task or also includes other tasks. Generic adaptation under an attribute-agnostic instruction $c_0$ and replay extend this approach to retaining category-level structure; conditional learning uses the grouping instruction $c$. Appendix~\ref{sec:training-path} lists the compared training paths.

\subsection{Instruction-Conditioned Supervision}
\label{sec:training-relations}

We define a \emph{contrastive unit} $U=(r,S)$ by the object scope $r$ and instructed attribute set $S$, and write $V_i^U=y_i^U$ for the attribute tuple defined above.

Each object--instruction pair is a conditional instance, indexed by $i,j$. Let $\mathcal B$ denote the conditional instances in the current physical batch on one device, with $|\mathcal B|=B$. For anchor $i\in\mathcal B$, define its positive set and within-task comparison set as
\begin{align}
    \mathcal P(i)&=\{j\in\mathcal B\setminus\{i\}:U_j=U_i,\;V_j^{U_i}=V_i^{U_i}\},\\
    \mathcal A_{\mathrm{within}}(i)&=\{j\in\mathcal B\setminus\{i\}:U_j=U_i\}.
\end{align}
The comparison set contains both positives and negatives; the within-task negative set is $\mathcal N_{\mathrm{intra}}(i)=\mathcal A_{\mathrm{within}}(i)\setminus\mathcal P(i)$. Within the same task, two distinct instances form a positive pair if their values agree on every instructed attribute; a difference in any instructed attribute makes the pair negative.

For conditional instance $i$, let $c_i$ be its instruction and write $\bm z_i=f_\theta(v_i,t_i,c_i)$, suppressing the instruction superscript for brevity. We use the normalized embedding $\bar{\bm z}_i=\bm z_i/\|\bm z_i\|_2$ and similarity $s_{ij}=\bar{\bm z}_i^\top\bar{\bm z}_j/\tau$, where $\tau>0$ is the temperature. InfoNCE~\citep{oord2018cpc} assigns one positive to each anchor; SupCon~\citep{khosla2020supervised} averages over multiple same-group instances in the batch. We adopt the latter, with the basic within-task objective
\begin{equation}
    \mathcal L_i^{\mathrm{within}}=-\frac{1}{|\mathcal P(i)|}
    \sum_{p\in\mathcal P(i)}
    \log\frac{\exp(s_{ip})}{\sum_{a\in\mathcal A_{\mathrm{within}}(i)}\exp(s_{ia})}.
    \label{eq:scoped-supcon}
\end{equation}
For each anchor, the loss averages the log-probability terms over all same-group positives. Appendix~\ref{app:gradient-analysis} derives the single-positive form and full gradients.

\subsection{Partition-Structured Sampling}

Random minibatches need not contain enough same-group examples to support a multi-positive objective, especially when a task has many groups. We therefore explicitly sample a local partition within each contrastive unit to provide same-group support and competing groups.

We sample a $C\times M$ structure per contrastive unit: $C$ distinct groups with $M\ge2$ distinct objects per group, sharing one instruction. For a complete unit, each anchor has $M-1$ positives and $M(C-1)$ within-task negatives, giving $CM-1$ comparison instances after excluding itself.

For compositional instructions, sampling classes are complete tuples $y_i^U$: positives agree on every target attribute, and a difference in any target attribute defines a within-unit negative. For task $u$, let $\mathcal G_u$ denote its available groups and $n_{u,g}$ the number of distinct objects in group $g$. A $C_u\times M_u$ unit requires
\[
\left|\{g\in\mathcal G_u:n_{u,g}\ge M_u\}\right|\ge C_u.
\]
We select $C_u$ groups that satisfy this support requirement and the label-separability filter, then sample $M_u$ distinct objects from each selected group. Batch construction is performed offline; each serialized record contains one complete batch.

\subsection{Heterogeneous Batches and Cross-Unit Contrast}

We combine heterogeneous $C\times M$ units at total batch size $B=20$, retaining each unit's instruction and labels. For example, C5M4 and C2M4+C3M4 both contain 20 instances but differ in whether negatives come only from the same task or also from other tasks. Appendix Table~\ref{tab:batch-schemas} lists exact schemas; CM24 controls separately use $B=24$.

A single forward pass encodes the $B$ conditional instances, and all pairwise comparisons reuse their representations. We consider two comparison scopes, denoted Within and Cross in the figures and tables. Within SupCon restricts comparisons to the anchor's contrastive unit. Cross SupCon retains these within-unit comparisons and additionally includes all instances from other units in the batch as negatives; its positive set is unchanged. The two objectives coincide for single-task batches; their effects on task-wise and pooled clustering are evaluated separately.

The Cross objective extends the comparison set from the anchor's unit to the entire batch:
\begin{equation}
    \mathcal A_{\mathrm{cross}}(i)
    =\mathcal A_{\mathrm{within}}(i)
    \cup\{j\in\mathcal B:U_j\neq U_i\}
    =\mathcal B\setminus\{i\}.
    \label{eq:full-comparison}
\end{equation}
The positive set $\mathcal P(i)$ is unchanged. The per-anchor loss is
\begin{equation}
    \mathcal L_i^{\mathrm{cross}}=
    -\frac{1}{|\mathcal P(i)|}
    \sum_{p\in\mathcal P(i)}
    \log\frac{\exp(s_{ip})}
    {\sum_{a\in\mathcal A_{\mathrm{cross}}(i)}\exp(s_{ia})}.
    \label{eq:cross-supcon}
\end{equation}

The training loss averages the per-anchor objective over valid instances in the physical batch; aggregation, loading-failure handling, and comparison counts appear in Appendix~\ref{app:batch-details}.

\section{Experiments}
\label{sec:experiments}

\subsection{Experimental Setup}
We evaluate three complementary targets: attribute grouping within each of 449 tasks, pooled clustering of 14,997 object--instruction instances, and generic clustering of 5,373 products by 50 leaf categories. Pooled reference groups retain task identities $(U,y_i^U)$ and do not merge products across types by shared attribute values. The 50 evaluation categories are excluded from our training sources. Embedding models receive common image--text fields and use full-native-dimensional vectors; HDBSCAN receives no ground-truth cluster count.

We compare general-purpose embeddings, adapted external conditional methods, and CIM-Embed training configurations. Direct configurations start from the official backbone; Within and Cross denote task-local and batch-wide contrastive comparisons. Staged configurations add attribute-agnostic adaptation before conditional training with generic replay. These paths are compared as complete configurations unless a control isolates an individual factor.

V-measure, ARI, coverage, and predicted cluster counts assess the recovered partitions; geometric statistics provide supporting diagnostics. V-measure and coverage are percentages. For these embedding comparisons, HDBSCAN parameters are selected using evaluation labels; the reported main-table scores are interpreted under this label-assisted selection protocol rather than as fixed-parameter held-out estimates. To assess sensitivity to this choice, we additionally evaluate fixed parameters, shared-grid selection, and retrospective five-fold leaf-grouped cross-validation. Direct Within retains a task-wise V-measure of 95.15 under leaf-disjoint parameter selection (Table~\ref{tab:external-unified-hdbscan}). External systems also differ in inputs and rejection policies. Appendix~\ref{app:evaluation-details} specifies metrics, aggregation, and comparison protocols; Appendix~\ref{app:validation-plan} reports supplementary controls and ablations. We first compare general embeddings and existing conditional systems, then examine training design, conditional--generic trade-offs, and transfer beyond product data.

Transfer evaluations use 8,029 Cards images, 10,531 CLEVR-4 synthetic scenes, and 50,000 CIFAR10-MC natural images, with two, four, and two grouping conditions, respectively. Encoders remain frozen on the target data. Cards uses a 12-point HDBSCAN grid; the other datasets each use a 48-point grid, selecting one shared parameter pair per model and dataset by mean conditional V. Target-pool labels are used for both selection and scoring, so these evaluations measure representation transfer under label-assisted clustering selection (Appendix~\ref{app:additional-transfer}).

\begin{table}[!htbp]
\centering
\caption{Task-wise conditional clustering and representation statistics, equally averaged over $449$ task units for general-purpose models, external methods, and direct and staged training configurations.}
\label{tab:embedding-main}
\begingroup
\scriptsize
\setlength{\tabcolsep}{2pt}
\renewcommand{\arraystretch}{1.10}
\begin{tabular*}{\linewidth}{@{\extracolsep{\fill}}l*{8}{r}@{}}
\toprule
Method & $V\uparrow$ & ARI$\uparrow$ & $Q\uparrow$ & Coverage$\uparrow$ & \shortstack{Clusters\\$\to 4.73$} & $R_{\mathrm{dist}}\downarrow$ & \shortstack{Align-\\ment$\downarrow$} & \shortstack{Unifor-\\mity$\downarrow$} \\
\midrule
\multicolumn{9}{@{}l}{\textbf{General-purpose models}} \\
Tongyi Flash & 45.62 & 0.1662 & 0.0462 & 73.14 & 7.57 & 0.9106 & 0.2533 & -0.5755 \\
Tongyi Plus & 47.45 & 0.1808 & 0.0483 & 74.41 & 7.76 & 0.8954 & 0.2455 & -0.5667 \\
Gemini Embedding 2 & 49.68 & 0.2044 & 0.0345 & 74.82 & 7.77 & 0.8756 & 0.2525 & -0.5929 \\
Qwen3-VL 2B (official) & 47.91 & 0.1882 & 0.0390 & 74.84 & 7.70 & 0.8911 & 0.5255 & -1.1854 \\
Qwen3-VL 8B (official) & 50.89 & 0.2185 & 0.0374 & 75.00 & 7.67 & 0.8634 & 0.5460 & -1.2665 \\
WeMM 2B (official) & 50.70 & 0.2120 & 0.0323 & 76.78 & 7.85 & 0.8733 & 0.5162 & -1.1896 \\
WeMM 9B (official) & 53.06 & 0.2418 & 0.0326 & 74.06 & 7.52 & 0.8519 & 0.4369 & -1.0329 \\
\midrule
\multicolumn{9}{@{}l}{\textbf{External methods}} \\
GCPM G1 & 57.32 & 0.3123 & 0.0454 & 76.06 & 6.99 & 0.7817 & 0.2116 & -0.5581 \\
GCPM G2 & 57.61 & 0.3083 & 0.0446 & 78.05 & 7.35 & 0.7797 & 0.3690 & -0.9585 \\
GCPM G3 & 74.85 & 0.5786 & 0.0454 & \sout{100.00} & 7.72 & 0.7817 & 0.2116 & -0.5581 \\
GCPM G4 & 74.17 & 0.5683 & 0.0446 & \sout{100.00} & 7.74 & 0.7797 & 0.3690 & -0.9585 \\
ABC & 54.52 & 0.3221 & 0.2142 & 86.80 & 6.22 & 0.7184 & 0.0915 & -0.3987 \\
CLAY & 51.86 & 0.2415 & 0.0526 & 85.99 & 8.34 & 0.8188 & 1.5644 & \textbf{-3.1463} \\
ITGC & 46.57 & 0.2103 & 0.1321 & 80.82 & 6.69 & 0.8613 & 0.0531 & -0.1564 \\
Multi-Sub & 30.21 & 0.0728 & 0.1643 & 75.42 & 4.92 & 1.1604 & 0.1755 & -0.2849 \\
\midrule
\multicolumn{9}{@{}l}{\textbf{Direct conditional training (Qwen3-VL 8B)}} \\
InfoNCE & 93.24 & 0.8943 & 0.4498 & 97.75 & 4.90 & 0.2430 & 0.2237 & -1.4857 \\
Within SupCon & \textbf{95.01} & \textbf{0.9238} & \textbf{0.5561} & 98.22 & 4.86 & \textbf{0.0972} & 0.0529 & -1.0779 \\
Cross SupCon & 94.68 & 0.9183 & 0.5551 & \textbf{98.50} & 4.87 & 0.1045 & 0.0590 & -1.0885 \\
\midrule
\multicolumn{9}{@{}l}{\textbf{Staged training (InfoNCE)}} \\
Stage 1 (generic) & 43.74 & 0.1738 & 0.1757 & 82.07 & 6.51 & 0.9780 & 0.0530 & -0.1318 \\
Stage 2 (generic) & 47.65 & 0.1950 & 0.1063 & 80.90 & 7.52 & 0.9027 & 0.2198 & -0.5208 \\
Stage 3 (instruction + replay) & 93.27 & 0.8933 & 0.4132 & 97.28 & 4.85 & 0.2904 & 0.2853 & -1.5887 \\
\midrule
\multicolumn{9}{@{}l}{\textbf{Staged training (SupCon)}} \\
Hierarchical SupCon (generic) & 43.85 & 0.1692 & 0.1651 & 82.32 & 6.87 & 0.9730 & \textbf{0.0401} & -0.0979 \\
Within + generic replay & 94.23 & 0.9104 & 0.5471 & 98.28 & \textbf{4.80} & 0.1116 & 0.0598 & -1.0779 \\
Cross + generic replay & 93.85 & 0.9054 & 0.5450 & 97.97 & 4.85 & 0.1162 & 0.0641 & -1.0905 \\
\bottomrule
\end{tabular*}
\endgroup
\par\smallskip
\begin{minipage}{\linewidth}\scriptsize
$V$ and coverage are percentages; ARI is unscaled. Scores retain noise label $-1$; cluster counts exclude noise. Geometric metrics are diagnostics. Struck-through GCPM coverage denotes complete assignment, not improved rejection. Embedding results retain evaluation-label-based selection. Full definitions and aggregation rules appear in Appendix~\ref{app:evaluation-details}.
\end{minipage}
\end{table}

\subsection{RQ1: Conditional Clustering with General-Purpose Embeddings}
\label{sec:rq1}

\paragraph{Conditional supervision makes instructed groups recoverable.}
General-purpose multimodal similarity does not adequately recover the attribute partitions evaluated here, whereas conditional training makes these partitions accessible to HDBSCAN. Direct Within SupCon reaches task-wise V-measure 95.01 versus 53.06 for WeMM 9B, the strongest evaluated general-purpose baseline; ARI rises from 0.2418 to 0.9238 while coverage increases from 74.06\% to 98.22\% (Table~\ref{tab:embedding-main}). Agreement with the target partition therefore improves while more objects receive non-noise assignments. The gain is not accompanied by a reduction in coverage to a small subset of easy objects.

\paragraph{Recorded no-instruction configurations yield lower partition recovery.}
Under the recorded no-instruction configuration, V is 43.16 for Direct Within and 42.35 for Direct InfoNCE, compared with 95.01 and 93.24 under their instructed configurations (Appendix Table~\ref{tab:historical-instruction-controls}). This large gap is consistent with a strong dependence of partition quality on the grouping condition. Because input templates and missing-data handling are not fully matched for these historical runs, we treat them as configuration-level controls rather than isolated instruction ablations (Appendix~\ref{sec:matched-pools}). High partition quality also extends to the evaluated two- and three-attribute tasks; independently sampled pools assess the reported compositions rather than unseen attribute combinations.

The tested dimensionality-reduction postprocessing does not close this gap; full UMAP results and protocols appear in Appendix~\ref{sec:dimension-controls}.

\subsection{RQ2: Comparison with Existing Conditional Clustering Methods}
\label{sec:rq2}

\paragraph{Direct embeddings provide an effective conditional clustering route.}
Across 449 tasks, Direct Within reaches V=95.01 and ARI=0.9238 from learned embeddings followed by HDBSCAN, compared with 74.85 and 0.5786 for GCPM G3, the strongest evaluated external configuration (Table~\ref{tab:embedding-main}). GCPM refinement raises V from 57.32/57.61 for G1/G2 to 74.85/74.17 for G3/G4, including reassignment of previously unclustered objects; its 100\% coverage reflects complete assignment rather than improved density-based rejection. ABC leads the remaining external methods at V=54.52, although CLAY exceeds ABC on three-attribute tasks (69.54 versus 65.86; Appendix Table~\ref{tab:external-dimensions}). Under the evaluated protocols, CIM-Embed achieves the highest aggregate partition agreement without intermediate description generation or language-model-guided merging. In the representation comparisons, the advantage persists under fixed, shared-grid, and leaf-disjoint HDBSCAN parameter selection (Table~\ref{tab:external-unified-hdbscan}). Inference-call statistics and system-comparison qualifications appear in Appendix~\ref{app:supporting-analyses}.

\subsection{RQ3: Contrastive Objectives and Batch Structure}
\label{sec:rq3}

\begin{table}[t]
\centering
\caption{Three evaluation scopes on Qwen3-VL 8B. V is a percentage; ARI and Q are unscaled. Task-wise and pooled Q use $Q_{\mathrm{target}}$; category Q uses $Q_{\mathrm{leaf}}$. Replay rows include prior generic adaptation. Full results appear in Table~\ref{tab:embedding-main} and Appendix Tables~\ref{tab:embedding-pooled} and~\ref{tab:embedding-generic}.}
\label{tab:scope-summary}
\begingroup\scriptsize
\setlength{\tabcolsep}{2pt}
\begin{tabular*}{\linewidth}{@{\extracolsep{\fill}}l*{9}{r}@{}}
\toprule
Configuration & \multicolumn{3}{c}{Task-wise} & \multicolumn{3}{c}{Pooled} & \multicolumn{3}{c}{Category} \\
\cmidrule(lr){2-4}\cmidrule(lr){5-7}\cmidrule(lr){8-10}
 & $V\uparrow$ & ARI$\uparrow$ & $Q\uparrow$ & $V\uparrow$ & ARI$\uparrow$ & $Q\uparrow$ & $V\uparrow$ & ARI$\uparrow$ & $Q\uparrow$ \\
\midrule
Qwen3-VL 8B & 50.89 & 0.2185 & 0.0374 & 59.21 & 0.0631 & 0.0364 & 89.32 & 0.6000 & 0.1132 \\
InfoNCE & 93.24 & 0.8943 & 0.4498 & 86.74 & 0.7919 & 0.2266 & 71.02 & 0.1423 & 0.0297 \\
Within & \textbf{95.01} & \textbf{0.9238} & \textbf{0.5561} & 91.85 & 0.8805 & 0.4279 & 75.46 & 0.2528 & 0.0618 \\
Cross & 94.68 & 0.9183 & 0.5551 & \textbf{93.86} & \textbf{0.9081} & \textbf{0.5207} & 85.02 & 0.5141 & 0.1704 \\
Within + replay & 94.23 & 0.9104 & 0.5471 & 90.85 & 0.8653 & 0.4021 & 96.96 & 0.8878 & 0.7430 \\
Cross + replay & 93.85 & 0.9054 & 0.5450 & 92.31 & 0.8856 & 0.4938 & \textbf{97.08} & \textbf{0.8879} & \textbf{0.7457} \\
\bottomrule
\end{tabular*}
\endgroup
\end{table}

\paragraph{Structured multi-positive training improves partition recovery across backbones.}
On Qwen3-VL 8B, InfoNCE, Within SupCon, and Cross SupCon obtain task-wise V=93.24, 95.01, and 94.68, respectively, while their pooled V-measures are 86.74, 91.85, and 93.86 (Tables~\ref{tab:embedding-main} and~\ref{tab:embedding-pooled}). The task-wise advantage of the structured SupCon recipe over InfoNCE also appears on Qwen 2B, WeMM 2B, and WeMM 9B (Appendix Table~\ref{tab:supplement-model-results}). The gain is therefore not specific to a single backbone. These comparisons evaluate the combined loss-and-sampling recipe.

\paragraph{Within and Cross favor different evaluation scopes.}
Within achieves the strongest task-wise result, whereas Cross increases pooled V-measure from 91.85 to 93.86 and reduces pooled $R_{\mathrm{dist}}$ from 0.2921 to 0.1940, with task-wise V changing from 95.01 to 94.68. Cross also improves pooled V-measure in both matched CM24 layouts (Appendix Table~\ref{tab:cm24-results-pooled}). The pooled partition scores indicate better preservation of task-specific groups under joint clustering, while the lower distance ratio provides complementary evidence of improved relative separation in the shared representation space. Together, these observations support the distinct local and pooled strengths of the two comparison scopes.

\paragraph{Batch structure should match the clustering objective.}
Among the seven Within configurations in the nine-run CM24 comparison, mixed C/M gives the highest task-wise V (93.17), while the single-/multi-Unit 5:5 mixture gives the highest pooled V (90.02; Appendix Tables~\ref{tab:cm24-results}--\ref{tab:cm24-results-generic}). There is consequently no single best layout across both targets, and neither more groups nor more positives guarantees improvement. Group coverage, positive support, and task mixing jointly define the training context; the supplementary configurations are reported in Appendix~\ref{sec:core-ablation}.

Qualitative UMAP examples for Wheel and Computer Case appear in Appendices~\ref{app:supporting-analyses} and~\ref{app:representation-visualization}.

\subsection{RQ4: Balancing Conditional and Generic Clustering}
\label{sec:rq4}

\paragraph{Category structure and conditional grouping are distinct capabilities.}
Generic adaptation produces strong category clustering with InfoNCE and hierarchical SupCon (V=95.88 and 96.21), but weak instructed attribute grouping (V=43.74 and 43.85; Tables~\ref{tab:embedding-main} and~\ref{tab:embedding-generic}). Conversely, direct conditional training improves instructed grouping while category V falls below that of the original Qwen3-VL 8B backbone. A representation can therefore preserve category-level structure without supporting instructed attribute partitions, and vice versa.

\paragraph{The staged configuration supports both structures.}
Relative to Direct Cross, staged Cross with generic replay raises category V from 85.02 to 97.08 while conditional V changes from 94.68 to 93.85. Staged Within with replay likewise combines category V=96.96 with conditional V=94.23. Both staged SupCon configurations therefore support strong category structure alongside high conditional partition quality. These results evaluate the complete staged configuration; replay and initialization effects are interpreted at the configuration level (Appendix~\ref{sec:training-path}).

\subsection{RQ5: Does Conditional Structure Transfer Beyond Product Data?}
\label{sec:rq5}

\paragraph{Representation transfer.}
Frozen product-trained encoders improve conditional grouping on Cards and CLEVR-4. Qwen 8B Direct Within reaches V=88.55 on Cards versus 56.36 for the strongest evaluated general-purpose baseline; the best evaluated CLEVR-4 configuration reaches 66.46 versus 42.07 for Qwen 8B Base. Learned grouping capability is therefore not confined to product data.

\paragraph{Transfer quality and instruction response are distinct.}
CLEVR-4 provides a matched test of instruction response on a fixed image pool: for each target attribute, images, reference labels, and selected clustering parameters are held fixed while only the instruction changes. Direct Within improves mean V by 56.58 percentage points over the no-instruction condition and by 59.22 percentage points averaged over alternative valid instructions. CIFAR10-MC shows representation improvements but weaker and less consistent instruction switching. Full results, controls, and configuration differences appear in Appendices~\ref{app:additional-transfer} and~\ref{app:transfer-diagnostics}. These experiments evaluate frozen encoder transfer under target-label-assisted clustering-parameter selection; Appendix~\ref{app:additional-transfer} details the selection protocol.

\section{Discussion and Conclusion}
\label{sec:limitations}
\label{sec:conclusion}

CIM-Embed trains multimodal representations with instruction-defined co-membership so that a standard clustering algorithm can recover the requested partition directly. The experiments distinguish within-task partition quality, cross-task organization, product-category structure, and instruction response on the same objects. Within supports strong local partitions, Cross repeatedly improves pooled results, and staged configurations balance conditional and category grouping; transfer experiments show that better partitions do not imply robust instruction switching. A conditional embedding should therefore be evaluated not only by whether it improves one target partition, but also by whether it organizes conditioned tasks globally, retains useful generic structure, and actually changes relations when the instruction changes.

The main ProductMIC scores use label-assisted HDBSCAN selection, while retrospective leaf-grouped cross-validation yields closely matching results, supporting robustness to leaf-disjoint parameter selection within the tested grid. The remaining limitations include noise in model-assisted annotations, target-label-assisted clustering selection in the cross-domain transfer evaluations, and the lack of independent isolation of loss and sampling contributions. Appendix~\ref{app:extended-limitations} details other data-coverage, training-path, and transfer boundaries. Future evaluation should further test transfer under independently selected clustering parameters and extend the controlled comparisons across training components.
\label{sec:main-end}

\ifdefined\CIMArxivVersion\else\clearpage\fi

\subsection*{REPRODUCIBILITY STATEMENT}
Sections~\ref{sec:benchmark} and~\ref{sec:method} specify the instruction-conditioned clustering task, model inputs, training objective, and structured batch construction. Section~\ref{sec:experiments} describes the evaluation framework and reports the available experimental results. Appendix~\ref{app:data-construction} documents ProductMIC construction, including source fields, deduplication, attribute normalization, label acceptance rules, task sampling, and the held-out category split. Appendix~\ref{app:gradient-analysis} provides the assumptions and derivations for the gradient analysis. Appendix~\ref{app:validation-plan} reports supplementary experimental results and their evaluation protocols. We will release the training code and the training and evaluation datasets upon acceptance, subject to applicable source terms and third-party rights.

\subsection*{ETHICS STATEMENT}
ProductMIC is constructed from product metadata in Amazon Reviews 2023. Neither training nor evaluation includes internal or non-public product data. The study focuses on grouping products from their images, titles, and descriptive text. Automatically generated attribute labels are filtered using consistency checks against normalized metadata and the acceptance rules in Appendix~\ref{app:data-construction}. These checks are intended to filter annotation errors but do not eliminate model bias or noisy supervision. Category coverage, missing attributes, and the use of small, balanced evaluation pools may underrepresent rare products and naturally occurring distributions; the reported results should not be assumed to establish reliability across other domains or deployment settings.

Public accessibility alone does not establish permission to redistribute every source image or text. Any release of derived resources should preserve source attribution, respect applicable source terms and third-party rights, and avoid exposing personal or confidential information. Downstream use requires attention to erroneous grouping and uneven performance across product categories, especially when clustering influences product visibility or access.

\subsection*{STATEMENT ON THE USE OF AI}
Qwen3.8-series models were used to extract product attributes during dataset construction. Model-assisted screening, metadata consistency checks, and supplementary-label acceptance rules are described in Appendix~\ref{app:data-construction}; accepted model-generated labels remain potentially noisy supervision. Generative AI also assisted with experimental-method review, code implementation and debugging, statistical-analysis scripts, interpretation of results, literature searches, and editing of manuscript text and structure. This included OpenAI Codex assistance with manuscript revision and LaTeX/PDF preparation.

AI does not replace experimental artifacts. Numerical results are recomputed from saved model outputs using fixed evaluation scripts. Important code changes are checked through tests or artifact-level verification. Causal and mechanism claims are restricted to evidence from completed controls. The authors are responsible for the text, code, numbers, citations, and conclusions in the final submission, and will review AI-assisted content and the corresponding submission disclosures before submission.

\clearpage
\bibliography{iclr2027_conference}\bibliographystyle{iclr2027_conference}

\clearpage\appendix
\section{ProductMIC Construction}
\label{app:data-construction}

Across the benchmark, task pools contain 10--98 products and 2--7 reference clusters.
\label{app:data-protocol}

\paragraph{Pipeline and scale.}
The ProductMIC pipeline comprises product deduplication and form selection, attribute-schema normalization, label verification, and task sampling for individual attributes and their combinations. This appendix documents the input fields, dataset statistics, and acceptance rules at each stage.

\subsection{Product Data and Attribute Annotation}
\label{sec:data-construction}
\label{sec:data-audit}

We construct ProductMIC from Amazon Reviews 2023 product metadata~\citep{hou2026bridging}, using parent ASIN identifiers for deduplication, product-form filtering, normalized attribute schemas, VLM annotation with metadata verification, and balanced task sampling. The following subsections specify these stages and their acceptance rules.

\subsection{Product Deduplication and Form Selection}

Products are identified by parent ASIN, and leaf categories are defined by complete category paths. Within a leaf category, products with identical normalized titles and main images are treated as duplicates, retaining the record with more complete information.

The form-discovery stage starts with 9,713 candidate leaf categories. We sample 32 products per category and use a VLM to identify their product forms. We retain categories in which the dominant form accounts for at least 70\% of sampled products, yielding 2,819 leaf categories. Subsequent product-level annotation checks again whether each object belongs to the target form.

\subsection{Attribute Schemas and Annotation Sampling}

For each selected leaf category, we select up to 20 raw attribute keys according to their product support in Details and audit their applicability to the target form, semantic clarity, and value types. Brand, product identifiers, rankings, reviews, and logistics fields are excluded as grouping attributes. Synonymous keys are merged within the same leaf category and product-form scope, and descriptions specify the referent and semantic scope of each attribute. Ambiguous attributes and non-boolean attributes with only one observed value are excluded from subsequent annotation; boolean attributes use a yes/no candidate domain.

Value normalization handles case, whitespace, punctuation, selected unit formats, and separators between letters, and merges synonymous expressions within the corresponding attribute. Candidate construction requires at least 18 supporting products per value within a leaf category, retains at most 7 candidate values per key, and requires at least two valid candidates.

We then jointly sample up to 128 products per leaf category, prioritizing coverage of different attribute values and increasing support for already covered value groups. The final annotation input contains 266,074 products, 11,357 attribute keys defined within category--form scopes, and 48,350 candidate values.

\subsection{Attribute Annotation and Label Acceptance}

The embedding input consists of the image, product title, and descriptive text assembled from description and features, plus the grouping instruction. It excludes Details, candidate attribute-value lists, and reference labels.

The VLM receives the product image, title, descriptive text assembled from description and features, target-form definition, attribute descriptions, and candidate values. It determines whether the product belongs to the target form and predicts values within the candidate domain. It may abstain when evidence is insufficient or the value falls outside that domain. The current product's Details field is withheld from the annotation model and used only for subsequent verification.

All accepted labels require that the product belongs to the target form, the attribute output is marked as a stable, valid value, and the value belongs to the corresponding candidate domain. Labels are then accepted under the following rules:
\begin{itemize}
    \item \textbf{Details-verified labels.}
    The predicted value agrees with the normalized value in the corresponding Details field. When multiple source keys are available, agreement with any valid source is sufficient.

    \item \textbf{Evidence-supplemented non-boolean labels.}
    The corresponding Details value is missing, product-form confidence is at least 0.9, the model returns exactly one judgment for the key, the evidence source includes the image, title, description, or features, and the evidence text is nonempty.

    \item \textbf{Supplementary boolean labels for missing values.}
    The corresponding Details value is missing, and the model returns a valid, stable boolean value. This branch does not impose the additional confidence and evidence-field requirements used for supplementary non-boolean labels.
\end{itemize}

A prediction that conflicts with an existing Details value cannot be accepted through a missing-value supplementation branch. Accepted labels undergo further synonym normalization within each leaf category, product form, and key, while retaining their provenance. Table~\ref{tab:label-provenance} reports the counts by source.

\begin{table}[!htbp]
\centering
\small
\caption{Label provenance. The training column counts the accepted-label pool after excluding evaluation leaf categories and before training sampling. Evaluation labels are deduplicated by product ID and key.}
\label{tab:label-provenance}
\begin{tabular}{lrr}
\toprule
Label source & Training pool & Evaluation \\
\midrule
Details-verified & 420,274 & 12,610 \\
Evidence-supplemented non-boolean & 106,498 & 952 \\
Supplementary boolean & 3,457 & 67 \\
\midrule
Total & 530,229 & 13,629 \\
\bottomrule
\end{tabular}
\end{table}

\subsection{Task Construction and Evaluation Split}
\label{sec:instruction-tasks}

\paragraph{Training-corpus and evaluation-resource statistics.}
Across the released InfoNCE and SupCon conditional-training formats, ProductMIC-Train covers 1,955 leaf categories, 163,168 distinct product IDs, and 474,458 distinct product--instruction pairs. These are union counts across the two formats, not the data seen by every model. The InfoNCE format contains 30,000 contrastive records covering 1,947 leaf categories, 132,255 distinct products, and 298,844 distinct product--instruction pairs. The SupCon format contains 18,000 batches of 20 products, covering 1,903 leaf categories, 132,255 distinct products, and 256,963 distinct product--instruction pairs. Both formats contain 360,000 product positions before repeated training exposure. Neither overlaps evaluation in leaf-category IDs or product IDs. These counts exclude generic adaptation and replay data. The accepted-label pool in Table~\ref{tab:label-provenance} precedes training sampling and is not a count of training records or exposures; ProductMIC-Bench statistics appear in Table~\ref{tab:eval-data-statistics}.

\paragraph{Evaluation attribute selection.}
The candidate audit data for the 50 evaluation leaf categories contain 6,226 products, 26,509 attribute labels, and 394 keys. The audit considers each attribute's meaning as a grouping criterion, the semantic level and mutual exclusivity of its values, alias normalization, and the verifiability of supporting evidence. Attributes are excluded if they mix semantic dimensions, have inconsistent granularity, or contain values that subsume one another.

For example, sealed/open and gas/radiant mix structure with energy source; smartphone/bar mixes product category with body form; and zoom versus zoom,wide angle exhibits value containment. Open-ended functionality, compatible-device, and package-content fields are also excluded when their value combinations are incomplete or their semantic levels are inconsistent. The audit retains 230 keys and 946 values and excludes 164 keys and 741 values.

Each instruction is constructed from the retained attribute names and descriptions. Only its target attributes require complete labels; missing values for other attributes do not exclude a product.

Following Section~\ref{sec:task-definition}, a task is $U=(r,S)$ and product $i$ has tuple $y_i^U$. Construction selects a pool $\mathcal X_U$ with complete target-attribute labels and implements co-membership as

\begin{equation}
    R_{ij}^U
    =\mathbb I[y_i^U=y_j^U]
    =\prod_{k\in S}\mathbb I[\ell_{ik}=\ell_{jk}],
    \qquad x_i,x_j\in\mathcal X_U.
    \label{eq:conditional-relation}
\end{equation}
Single-, two-, and three-attribute tasks correspond to $|S|=1,2,3$, respectively.

\paragraph{Single-attribute tasks.}
Each attribute value must be supported by at least 5 distinct products, and each attribute must retain at least two valid values. The number sampled per class is
\begin{equation}
    M_U=\min\left(20,\min_{v\in\mathcal V_U}n_v\right),
\end{equation}
where $\mathcal V_U$ is the task's set of target values and $n_v$ is the number of products supporting value $v$. Products are selected after sorting by a stable hash. The resulting single-attribute tasks contain 2--7 classes and 5--20 products per class.

\paragraph{Composite-attribute tasks.}
Two- and three-attribute tasks require complete labels for all constituent attributes. For each combination, we retain up to 6 composite classes with the highest product support, subject to the following conditions:
\begin{itemize}
    \item At least 4 composite classes, each with at least 4 products;
    \item At least two marginal values for each constituent attribute;
    \item Maximum pairwise NMI among constituent attributes no greater than 0.60.
\end{itemize}
We then sample equal numbers of products across composite classes, with at most 20 products per class. All 50 leaf categories support eligible two-attribute tasks, and 14 support eligible three-attribute tasks. All tasks passing these filters are included in the benchmark.

\paragraph{Final benchmark statistics.}
The final evaluation set contains $5{,}373$ deduplicated products, $449$ task units, and $14{,}997$ product--task records with instructions, as summarized in Table~\ref{tab:eval-data-statistics}. Single-attribute and two-attribute tasks cover all $50$ evaluation leaf categories; three-attribute tasks cover the $14$ categories that satisfy the selection criteria. A product may participate in multiple tasks, so product counts overlap across attribute orders. Each task contains $10$ to $98$ products and $2$ to $7$ ground-truth clusters, supporting evaluation of fine-grained grouping in small, controlled product pools.

\begin{table}[!htbp]
\centering
\small
\caption{Evaluation-set and task-grouping statistics. Products are deduplicated by ID, with overlap across attribute orders. Product and record counts include only the instructed condition. The final three rows report minimum/median/maximum values; cluster sizes are computed over ground-truth clusters within tasks.}
\label{tab:eval-data-statistics}
\label{tab:data-audit}
\setlength{\tabcolsep}{5pt}
\begin{tabular}{lrrrr}
\toprule
Statistic & \shortstack{One\\attribute} & \shortstack{Two\\attributes} & \shortstack{Three\\attributes} & Overall \\
\midrule
Leaf categories & 50 & 50 & 14 & 50 \\
Task units & 230 & 189 & 30 & 449 \\
Distinct products & 5,053 & 3,065 & 512 & 5,373 \\
Product--task records & 9,105 & 5,211 & 681 & 14,997 \\
\midrule
Products per task & 10/40/98 & 16/24/72 & 16/22/30 & 10/30/98 \\
Ground-truth clusters/task ($K$) & 2/4/7 & 4/6/6 & 4/5/6 & 2/5/7 \\
Ground-truth cluster size & 5/8/20 & 4/5/18 & 4/4/7 & 4/6/20 \\
\bottomrule
\end{tabular}
\end{table}

\paragraph{Held-out split.}
\label{sec:data-splits}
All 50 evaluation leaf categories are excluded from the training sources. Training uses the remaining categories, so this split evaluates leaf categories unseen during our training stage.

\subsection{Instruction Controls}
\label{sec:matched-pools}

\paragraph{Within-task instruction controls.}
For WeMM 9B Direct Cross, the no-instruction condition preserves the same image and product text and omits only the grouping instruction; all 14,997 instructed records add that instruction as a prefix. For other recorded no-instruction configurations, input templates and missing-data handling are not guaranteed to be identical, so we treat them as configuration-level controls rather than matched instruction ablations (Table~\ref{tab:historical-instruction-controls}). A fixed generic instruction $c_0$ is a separate condition. A matched ablation also holds preprocessing and clustering parameters fixed and scores both conditions against the task target; category evaluation in Table~\ref{tab:embedding-generic} uses different labels and cannot substitute for it.

\paragraph{Cross-instruction controls.}
Independent class filtering and sampling can produce different pools across tasks. A matched cross-instruction test therefore requires the intersection of the evaluated pools within a product-form scope, fixed images, text, preprocessing, and clustering parameters, and separate embeddings under each instruction. The retained fraction of each original pool characterizes the test coverage. Task pairs with identical target partitions can assess quality but cannot demonstrate relation switching. We therefore do not use the current ProductMIC results as evidence of exact relation switching on a matched object pool.

\section{Training Details}
\label{app:batch-details}

Let $\mathcal U(\mathcal B)$ be the set of units in a batch and $N_u=C_uM_u$ the number of instances in unit $u$. For complete batches in which every instance has at least one positive, we average within each unit and aggregate in proportion to its size:
\begin{align}
    \mathcal L_u
    &=\frac{1}{N_u}
      \sum_{\substack{i\in\mathcal B\\U_i=u}}
      \mathcal L_i^{\mathrm{cross}},
    \label{eq:unit-supcon}\\
    \mathcal L_{\mathrm{train}}
    &=\frac{1}{B}
      \sum_{u\in\mathcal U(\mathcal B)}N_u\mathcal L_u
      =\frac{1}{B}
      \sum_{i\in\mathcal B}\mathcal L_i^{\mathrm{cross}}.
    \label{eq:final-objective}
\end{align}
All valid anchors therefore receive equal loss weight. If loading failures remove instances, the implementation averages over the remaining anchors with at least one positive and a finite contrastive denominator. Appendix~\ref{app:gradient-analysis} derives the exact relation between Cross and the within-task Within objective: expanding the comparison set introduces cross-unit terms and redistributes within-unit softmax weights.

For an anchor in a unit of size $N_u$, the cross-unit negative count and total candidate count are
\[
    N_{\mathrm{cross}}=B-N_u,\qquad
    D_{\mathrm{full}}=B-1.
\]
At fixed $B$ and $M$, configurations preserve positive/negative counts while changing negatives' task sources. C5M4 and C2M4+C3M4 both provide 3 positives and 16 negatives; the latter replaces some within-unit comparisons with cross-unit comparisons.

\begin{table}[!htbp]
\centering
\caption{Per-anchor comparison counts for four complete batch configurations at $B=20$ under Cross SupCon. $P$ counts positives; $N_{\mathrm{intra}}$ and $N_{\mathrm{cross}}$ count within-unit and cross-unit negatives; $D_{\mathrm{full}}$ is the total candidate count. Slashes separate the counts for the two units.}
\label{tab:batch-schemas}
\begin{tabular}{lcccc}
\toprule
Schema & $P$ & $N_{\mathrm{intra}}$ & $N_{\mathrm{cross}}$ & $D_{\mathrm{full}}$ \\
\midrule
C5M4 & 3 & 16 & 0 & 19 \\
C4M5 & 4 & 15 & 0 & 19 \\
C2M4+C3M4 & 3 / 3 & 4 / 8 & 12 / 8 & 19 / 19 \\
C2M5+C2M5 & 4 / 4 & 5 / 5 & 10 / 10 & 19 / 19 \\
\bottomrule
\end{tabular}
\end{table}

\paragraph{Initialization and replay.}
\label{sec:training-path}
The completed direct and staged configurations are reported in Tables~\ref{tab:embedding-main}, \ref{tab:embedding-pooled}, and~\ref{tab:embedding-generic}. These path comparisons do not independently isolate initialization and replay effects.

\subsection{Scope of Configuration Comparisons}
Direct training starts from the corresponding official backbone, whereas staged training adds generic adaptation before conditional training with replay. Because these comparisons jointly vary initialization history, training objective, and replay rather than matching all training conditions and budgets, staged results are interpreted as configuration-level comparisons, not isolated ablations of replay or loss effects. Tables distinguish the earlier InfoNCE replay checkpoint from Stage 3 InfoNCE with replay.
\FloatBarrier

\section{Gradient Analysis}
\label{app:gradient-analysis}

This appendix analyzes the contrastive objectives in Section~\ref{sec:method}, clarifying the respective effects of structured sampling and comparisons across contrastive units. We use the standard InfoNCE and SupCon formulations~\citep{oord2018cpc,khosla2020supervised} to derive the exact relation between Within and Cross, the full bidirectional gradients, and backpropagation through normalization.

\subsection{Unified Objectives and Structured Sampling}
\label{app:gradient-notation}

Following the main text, the unnormalized representation of conditional instance $i$ is $\bm z_i=f_\theta(x_i,c_i)$, its normalized representation is $\bar{\bm z}_i=\bm z_i/\|\bm z_i\|_2$, and the logit is $s_{ij}=\bar{\bm z}_i^\top\bar{\bm z}_j/\tau$. We assume $\bm z_i\neq\bm0$, $\tau>0$, $\varnothing\neq\mathcal P(i)\subseteq\mathcal A(i)$, and $i\notin\mathcal A(i)$. The full gradients correspond to the equally weighted average over instances in Equation~\eqref{eq:final-objective}: every instance serves as an anchor, and gradients propagate through all candidate representations. The formulas must be adjusted to the actual computational graph if candidate gradients are stopped, the comparison domain includes other devices, or the loss uses different weights.

Let $K_i=|\mathcal P(i)|$ and $y_{ij}=\mathbb I[j\in\mathcal P(i)]/K_i$. For a chosen candidate domain $\mathcal A(i)$, the per-anchor objective and its logit derivative can be written uniformly as
\begin{align}
 q_{ij}&=\frac{\exp(s_{ij})}{\sum_{a\in\mathcal A(i)}\exp(s_{ia})},\qquad j\in\mathcal A(i),\label{eq:appendix-softmax}\\
 \mathcal L_i&=-\sum_{p\in\mathcal P(i)}\frac{s_{ip}}{K_i}
       +\log\sum_{a\in\mathcal A(i)}\exp(s_{ia}),\qquad
 \frac{\partial\mathcal L_i}{\partial s_{ij}}=q_{ij}-y_{ij}.
 \label{eq:appendix-supcon-logit-grad}
\end{align}
We set $q_{ij}=y_{ij}=0$ outside the candidate domain. This recovers single-positive InfoNCE when $K_i=1$ and multi-positive SupCon when $K_i>1$. Because $s_{ij}$ already includes the temperature, the logit derivative requires no additional factor of $1/\tau$.

For a single anchor loss, the derivative with respect to its normalized representation is
\begin{equation}
 \nabla_{\bar{\bm z}_i}\mathcal L_i
 =\frac1\tau\left(\sum_{j\in\mathcal A(i)}q_{ij}\bar{\bm z}_j
              -\frac1{K_i}\sum_{p\in\mathcal P(i)}\bar{\bm z}_p\right).
 \label{eq:appendix-supcon-vector-grad}
\end{equation}
The second term uses the local same-group mean in the current batch. The logit derivative for a positive pair is $q_{ip}-1/K_i$, which can be positive when $q_{ip}>1/K_i$; SupCon therefore does not require similarity to increase for every positive pair at every step. This expression is only the derivative of $\mathcal L_i$ with respect to its anchor and does not yet include the gradients arising when instance $i$ serves as a candidate for other anchors.

Under $C_u\times M_u$ sampling for contrastive unit $u$, each attribute tuple has $M_u\geq2$ instances, so $K_i=M_{U_i}-1$ and the number of within-task negatives is $M_{U_i}(C_{U_i}-1)$. Structured sampling retains the same single-batch SupCon form but changes positive support, class coverage, and the batch distribution, and can therefore change the expected training objective. For a fixed encoder, contrastive unit, and attribute tuple, if same-group instances are approximately independent and identically distributed with covariance $\Sigma_{u,V}$, the covariance of the local positive mean is $\Sigma_{u,V}/K_i$; correlated samples or sampling without replacement require corrections. This relation describes mean estimation and does not directly imply lower variance of the full gradient or improved final clustering.

\subsection{Exact Relation between Within and Cross}
\label{app:gradient-cross-unit}

A physical batch $\mathcal B$ contains multiple contrastive units and has size $B=\sum_u C_uM_u$. Positives remain restricted to instances with the same contrastive unit and complete attribute tuple. Let $X\in\{\mathrm W,\mathrm C\}$ denote Within and Cross, respectively:
\begin{equation}
 \mathcal A_{\mathrm W}(i)=\{j\in\mathcal B\setminus\{i\}:U_j=U_i\},\qquad
 \mathcal A_{\mathrm C}(i)=\mathcal B\setminus\{i\}.
 \label{eq:appendix-comparison-domains}
\end{equation}
Substituting the corresponding candidate domain into Equation~\eqref{eq:appendix-softmax} yields $q^X$ and $\mathcal L_i^X$. The logit derivative for a cross-unit pair is zero under Within and $q_{ij}^{\mathrm C}>0$ under Cross, introducing a cross-unit repulsion coefficient at the level of independent logits. Because representations and parameters are shared across pairs, this sign does not guarantee that all cross-unit similarities decrease after a network update.

For the same current representations and temperature, define
\begin{align}
 Z_{\mathrm{intra}}(i)&=\sum_{a\in\mathcal B\setminus\{i\},\,U_a=U_i}\exp(s_{ia}),&
 Z_{\mathrm{cross}}(i)&=\sum_{a\in\mathcal B:\,U_a\neq U_i}\exp(s_{ia}),\\
 \alpha_i&=\frac{Z_{\mathrm{intra}}(i)}{Z_{\mathrm{intra}}(i)+Z_{\mathrm{cross}}(i)}\in(0,1],&
 q_{ij}^{\mathrm C}&=\alpha_iq_{ij}^{\mathrm W}\quad(U_j=U_i,\ j\neq i).
 \label{eq:appendix-alpha}
\end{align}
Here, $\alpha_i$ is the total probability mass assigned to within-unit candidates under Cross. It equals $1$ when no cross-unit candidates are present and is strictly less than $1$ when cross-unit candidates exist and their logits are finite.

Because the two objectives have identical positive terms, they admit the exact decomposition
\begin{equation}
 \boxed{\mathcal L_i^{\mathrm C}
 =\mathcal L_i^{\mathrm W}+\log\left(1+\frac{Z_{\mathrm{cross}}(i)}{Z_{\mathrm{intra}}(i)}\right)
 =\mathcal L_i^{\mathrm W}-\log\alpha_i.}
 \label{eq:appendix-loss-decomposition}
\end{equation}
Thus, Cross equals Within plus a competition penalty that encourages greater total probability mass for within-unit candidates. Its gradient changes both within-unit and cross-unit coefficients; it cannot be interpreted as leaving the original within-unit gradient unchanged and merely adding cross-unit repulsion.

Specifically, let $g_{ij}^X=\partial\mathcal L_i^X/\partial s_{ij}$. Then
\begin{align}
 g_{ip}^{\mathrm C}-g_{ip}^{\mathrm W}
   &=(\alpha_i-1)q_{ip}^{\mathrm W}\leq0,&&p\in\mathcal P(i),\\
 g_{in}^{\mathrm C}
   &=\alpha_i g_{in}^{\mathrm W},&&U_n=U_i,\ n\notin\mathcal P(i),\ n\neq i,\\
 g_{ik}^{\mathrm C}&=q_{ik}^{\mathrm C},\qquad
 \sum_{k:\,U_k\neq U_i}q_{ik}^{\mathrm C}=1-\alpha_i,&&U_k\neq U_i.
 \label{eq:appendix-mass-redistribution}
\end{align}
At fixed representations, Cross shifts positive-pair coefficients toward attraction, dilutes the coefficient of each within-unit negative pair, and assigns nonzero coefficients to cross-unit pairs. Positive-pair coefficients need not become negative. The quantity $1-\alpha_i$ is the total mass of cross-unit logit coefficients, not the norm of the full representation gradient.

For Cross to approximate Within, $Z_{\mathrm{cross}}(i)/Z_{\mathrm{intra}}(i)\ll1$ is required. This ratio depends jointly on similarities, candidate counts, and temperature; separation between contrastive units alone does not establish that it approaches zero.

\subsection{Full Bidirectional Gradients and Normalization}
\label{app:gradient-full-vector}

Let $\mathcal L^X=B^{-1}\sum_{i\in\mathcal B}\mathcal L_i^X$. An instance $r$ serves both as an anchor and as a candidate for other anchors. Summing these contributions gives
\begin{equation}
 \boxed{\nabla_{\bar{\bm z}_r}\mathcal L^X
 =\frac1{B\tau}\sum_{k\neq r}
 \left(q_{rk}^X+q_{kr}^X-y_{rk}-y_{kr}\right)\bar{\bm z}_k.}
 \label{eq:appendix-full-vector}
\end{equation}
When each group within a contrastive unit contains $M_u$ sampled instances, every positive pair satisfies $K_r=K_k=M_u-1$ and has coefficient $q_{rk}^X+q_{kr}^X-2/(M_u-1)$; a negative pair has coefficient $q_{rk}^X+q_{kr}^X$. Cross-unit coefficients are zero under Within, while Cross retains the corresponding bidirectional terms. The value of $M_u$ may differ across contrastive units. Even when $s_{rk}=s_{kr}$, different anchors can have different denominators, so in general $q_{rk}^X\neq q_{kr}^X$.

Differentiating Equation~\eqref{eq:appendix-loss-decomposition} also gives
\begin{equation}
 \nabla_{\bar{\bm z}_r}\mathcal L^{\mathrm C}
 =\nabla_{\bar{\bm z}_r}\mathcal L^{\mathrm W}
  +\nabla_{\bar{\bm z}_r}\left[-\frac1B\sum_{i\in\mathcal B}\log\alpha_i\right].
 \label{eq:appendix-gradient-decomposition}
\end{equation}
The penalty gradient requires differentiating $\alpha_i$ for every relevant anchor. Vector contributions can cancel, so the magnitude of an individual logit coefficient is neither the magnitude of the full update nor a direct determinant of the compactness of the final local partition.

For the actual network output $\bm z_r$, the Jacobian of $L_2$ normalization and the corresponding backpropagation are
\begin{equation}
 J_r=\frac{\partial\bar{\bm z}_r}{\partial\bm z_r}
   =\frac{\bm I-\bar{\bm z}_r\bar{\bm z}_r^\top}{\|\bm z_r\|_2},\qquad
 \nabla_{\bm z_r}\mathcal L^X=J_r\nabla_{\bar{\bm z}_r}\mathcal L^X.
 \label{eq:appendix-normalized-grad}
\end{equation}
The matrix $J_r$ projects the gradient onto the tangent space of the unit sphere and scales it by the output norm. Updating encoder parameters further requires propagation through $\partial\bm z_r/\partial\theta$. Accordingly, ``attraction'' and ``repulsion'' here primarily describe the objective's local preferences over similarities and do not replace an analysis of full parameter updates or the geometry after convergence.

\subsection{Relation to Empirical Observations and Its Limits}
\label{app:gradient-experimental-support}

In the direct instruction-training configurations reported in the main text, Within and Cross SupCon obtain task-wise $R_{\mathrm{dist}}$ values of $0.0972$ and $0.1045$, pooled $R_{\mathrm{dist}}$ values of $0.2921$ and $0.1940$, and Uniformity values of $-2.5084$ and $-3.2891$, respectively. These observations show that, in this comparison, the pooled relative distance ratio decreases while the task-wise ratio slightly increases; the Uniformity change describes the distribution geometry~\citep{wang2020alignmentuniformity}. Cross-unit comparisons do not necessarily improve within-task clustering, as reflected by the different task-wise and pooled trends.

Treating all cross-unit instances as negatives is a supervision design choice and does not imply that the semantics under different instructions are necessarily mutually exclusive. If pooled ground truth is jointly defined by $(U,V)$, stronger separation by contrastive-unit identity alone may improve some pooled metrics without demonstrating correct instruction-dependent changes in relations on the same product pool. Evaluation must therefore examine within-task grouping, conditional changes on a shared object pool, and pooled performance together. These derivations also do not guarantee better generic clustering. The role of $C\times M$ is to organize supervision and sampling, rather than to introduce a new fundamental SupCon loss.

\FloatBarrier
\FloatBarrier

\section{Evaluation Protocol}
\label{app:evaluation-details}

\subsection{Evaluation Scopes and Metric Definitions}

Embedding baselines and our models share evaluation objects and image--text fields; HDBSCAN receives no ground-truth cluster count. Task-wise evaluation averages 449 tasks equally. Pooled evaluation clusters all object--instruction instances while retaining task-specific group identities $(U,y_i^U)$, measuring their separation rather than cross-type merging by shared attribute values. Generic evaluation targets 50 leaf categories. Embeddings use full native dimensionality without MRL. V-measure, ARI, coverage, and cluster counts assess partitions; geometric statistics provide supporting diagnostics. Training-corpus statistics and the audited no-instruction input construction appear in Appendix~\ref{app:data-construction}; limits of configuration-level attribution appear in Appendix~\ref{app:extended-limitations}.

Baselines include Qwen, WeMM, and Gemini Embedding 2. We compare direct conditional training with generic adaptation followed by conditional training and generic replay. Data, initialization, positive/negative sources, physical batches, and budgets must be distinguished before attributing path differences to individual losses or replay.

Notes for all three tables: $V$ and coverage are percentages. Task-wise and pooled evaluations use $Q=Q_{\mathrm{target}}$; generic evaluation uses $Q=Q_{\mathrm{leaf}}$, all with $T=1$ (Equation~\eqref{eq:q-recovery}). $R_{\mathrm{dist}}$, Alignment, and Uniformity describe representation geometry rather than clustering quality. $\uparrow/\downarrow$ indicate preferred directions; Clusters is the mean predicted count for task-wise evaluation (mean ground truth 4.73), and the global count for pooled/generic evaluation (ground truth 2,123/50). Best reported values, including ties, are bold. Embeddings use full native dimensionality; our trained models do not use MRL.
ARI uses its original scale in all three tables; -- denotes an unreported metric. GCPM G1 uses proxy-text representations and G2 uses original image--text plus proxies with fixed HDBSCAN; G3/G4 add semantic merging to G1/G2, respectively. G1--G4, ABC, CLAY, ITGC, and Multi-Sub each completed all 449 task units and 14,997 records. Scores are unit-macro averages; noise label $-1$ is retained as one predicted cluster, and coverage is reported separately.

For embedding baselines and our models, task-wise V-measure and ARI are averaged equally over 449 tasks. Pooled evaluation first clusters all 14,997 object--instruction instances jointly, then computes V-measure and ARI within each original task and averages them equally. These scores measure the preservation of task-specific partitions under joint clustering. Generic V-measure and ARI are computed over 5,373 products using the 50 leaf-category labels. All three scopes retain noise label $-1$ as one predicted cluster.Missing task-wise predictions were recovered from stored embeddings with the original selected parameters, without a new search; per-task V-measure and coverage and mean cluster counts reproduce the original results. This metric backfill retains the evaluation-label-based parameter selection protocol.
Struck-through coverage marks GCPM G3/G4 as complete-assignment sorting systems: their 100\% coverage follows from this output protocol, does not establish better density-based rejection, and is excluded from coverage best-value comparisons. G1/G2 and direct HDBSCAN methods may retain noise; their coverage describes the fraction assigned to non-noise clusters.
Alignment averages distinct same-class squared distances between L2-normalized vectors equally over classes, then equally over original tasks in both task-wise and pooled evaluation, so their Alignment values coincide. Uniformity uses $\log\operatorname{mean}_{i\ne j}\exp(-2\|z_i-z_j\|_2^2)$: task-wise values are averaged over the 449 within-task evaluations; pooled values use all 14,997 conditional records; generic values use the complete 5,373-product pool. All geometric metrics retain noise objects. Generic $R_{\mathrm{dist}}$ is the ratio of class-macro within-class and nearest-other-class cosine distances over the 50 leaf categories (one evaluation pool).
InfoNCE Stages 2 and 3 use non-MRL checkpoints: Stage 2 uses generic training; Stage 3 uses conditional training with generic replay and differs from the earlier InfoNCE replay checkpoint.

\paragraph{Density-hierarchy recovery score.}
\label{sec:q-metric}
On L2-normalized embeddings, cosine distances define core distances to the fifth other neighbor. Mutual-reachability distances are the maximum of the pairwise distance and the two core distances. A minimum spanning tree induces components as the distance threshold increases, merging tied edges simultaneously. For each non-singleton, non-root component $A$, let $a_A,b_A$ be its birth and merge thresholds and $\ell_A=\log(b_A/a_A)$ its positive lifetime (infinite when $a_A=0$). For eligible reference groups $\mathcal G$ with at least two objects and smaller than the full pool, the recovery statistic is
\begin{equation}
 Q_T=\frac{1}{T}\int_0^T\frac{1}{|\mathcal G|}\sum_{G\in\mathcal G}
 \max_{A:\ell_A>t}\frac{|G\cap A|}{|G\cup A|}\,dt,
 \qquad T=1.
 \label{eq:q-recovery}
\end{equation}
The maximum is zero when no component qualifies. The statistic averages best Jaccard recovery across reference groups and integrates over component lifetime; it is not a score of the final HDBSCAN partition. Task-wise and pooled targets use task-specific groups ($Q_{\mathrm{target}}$), while category evaluation uses leaf labels ($Q_{\mathrm{leaf}}$). Task-wise scores average equally over tasks; pooled hierarchy construction uses the full pool before aggregation by original task.

\paragraph{Relative distance metric.}
We replace the two absolute-distance columns with a relative distance ratio:
\begin{equation}
R_{\mathrm{dist}}=\frac{1}{|\mathcal U|}\sum_{u\in\mathcal U}
\frac{D_{\mathrm{in}}^{(u)}}{D_{\mathrm{out}}^{(u)}}.
\label{eq:distance-ratio}
\end{equation}
Within each task, $D_{\mathrm{in}}^{(u)}$ averages cosine distances between distinct same-class objects equally over ground-truth classes. $D_{\mathrm{out}}^{(u)}$ averages each class's mean distance to its nearest other class, also equally over classes. We divide these task-level quantities before averaging equally over all 449 tasks. Task-wise evaluation searches for other classes within the current task; pooled evaluation defines classes by $(U,V)$ and searches the entire pool, then aggregates by original task. The ratio is computed before task averaging, can exceed one, and is a geometric diagnostic rather than a partition-quality score. All reported denominators are positive. Alignment retains its original definition.
\FloatBarrier

\section{Additional ProductMIC Results}
\label{app:validation-plan}

This appendix reports ProductMIC instruction, dimensionality, batch-composition, backbone, and category-retention results.

\paragraph{Dimensionality, postprocessing, and instruction controls.}
\label{sec:dimension-controls}
Full-dimensional and UMAP results appear in Tables~\ref{tab:umap-task}--\ref{tab:umap-generic}. Recorded prefix, instruction, and attribute-order comparisons appear in Table~\ref{tab:historical-instruction-controls}. Different attribute orders use different task pools and do not isolate compositional difficulty.

Tables~\ref{tab:umap-task}--\ref{tab:umap-generic} report the completed fixed-postprocessing comparison for 12 models and three evaluation scopes. Full-dimensional vectors are L2-normalized before UMAP (5 components, 5 neighbors, minimum distance 0, Euclidean metric, seed 42), with no normalization after projection. Euclidean EOM HDBSCAN reuses the previously selected minimum cluster size and minimum samples for each model and scope, without another search. This measures postprocessing with the selected parameters, not jointly optimized UMAP and HDBSCAN. Parameter selection did not consistently use independent validation data, so this is not a held-out evaluation under a frozen validation protocol.

UMAP is fitted independently within each task, once over all 14,997 object--instruction records for pooled evaluation, and once over 5,373 objects for generic evaluation, without passing labels to UMAP. Task-wise and pooled V, ARI, and coverage are macro-averaged over 449 tasks; generic scores are global. Cluster counts exclude noise and are averaged across tasks only in the task-wise scope; scoring retains the common noise label $-1$. V and coverage are percentages; ARI is unscaled. Flash/Plus are excluded from these UMAP tables because their input-fusion paths are not established by the available records. Recorded 64-/128-dimensional prefix results are available in Table~\ref{tab:historical-instruction-controls}.

\begin{table}[!htbp]
\centering
\caption{UMAP comparison: task-wise clustering. Each cell reports full dimensions $\to$ UMAP.}
\label{tab:umap-task}
\begingroup
\scriptsize
\setlength{\tabcolsep}{2pt}
\resizebox{\linewidth}{!}{%
\begin{tabular}{@{}lrrrr@{}}
\toprule
Model & $V$ & ARI & Coverage & Clusters \\
\midrule
Gemini Embedding 2 & 49.68$\to$53.80 & 0.2044$\to$0.2796 & 74.82$\to$97.92 & 7.77$\to$8.11 \\
Qwen3-VL 2B (official) & 47.91$\to$51.99 & 0.1882$\to$0.2471 & 74.84$\to$97.86 & 7.70$\to$8.62 \\
Qwen3-VL 8B (official) & 50.89$\to$56.59 & 0.2185$\to$0.3085 & 75.00$\to$97.75 & 7.67$\to$8.31 \\
WeMM 2B (official) & 50.70$\to$54.70 & 0.2120$\to$0.2794 & 76.78$\to$98.02 & 7.85$\to$8.43 \\
WeMM 9B (official) & 53.06$\to$60.53 & 0.2418$\to$0.3575 & 74.06$\to$97.93 & 7.52$\to$8.25 \\
Direct InfoNCE & 93.24$\to$92.64 & 0.8943$\to$0.8824 & 97.75$\to$99.93 & 4.90$\to$5.21 \\
Direct Within & 95.01$\to$92.97 & 0.9238$\to$0.8913 & 98.22$\to$99.94 & 4.86$\to$5.24 \\
Direct Cross & 94.68$\to$92.35 & 0.9183$\to$0.8820 & 98.50$\to$99.97 & 4.87$\to$5.42 \\
Stage1 InfoNCE & 43.74$\to$47.30 & 0.1738$\to$0.2154 & 82.07$\to$98.20 & 6.51$\to$7.89 \\
No-instruction Hier & 43.85$\to$46.78 & 0.1692$\to$0.2070 & 82.32$\to$98.04 & 6.87$\to$7.87 \\
Within + Replay & 94.23$\to$92.94 & 0.9104$\to$0.8903 & 98.28$\to$99.86 & 4.80$\to$5.15 \\
Cross + Replay & 93.85$\to$92.49 & 0.9054$\to$0.8832 & 97.97$\to$99.93 & 4.85$\to$5.21 \\
\bottomrule
\end{tabular}}
\endgroup
\end{table}

\begin{table}[!htbp]
\centering
\caption{UMAP comparison: pooled conditional clustering. Each cell reports full dimensions $\to$ UMAP.}
\label{tab:umap-pooled}
\begingroup
\scriptsize
\setlength{\tabcolsep}{2pt}
\resizebox{\linewidth}{!}{%
\begin{tabular}{@{}lrrrr@{}}
\toprule
Model & $V$ & ARI & Coverage & Clusters \\
\midrule
Gemini Embedding 2 & 59.04$\to$57.86 & 0.0417$\to$0.0511 & 94.48$\to$90.48 & 3806$\to$3625 \\
Qwen3-VL 2B (official) & 59.37$\to$58.35 & 0.0354$\to$0.0493 & 94.60$\to$91.72 & 4164$\to$3681 \\
Qwen3-VL 8B (official) & 59.21$\to$57.70 & 0.0631$\to$0.0692 & 91.59$\to$88.45 & 4002$\to$3625 \\
WeMM 2B (official) & 59.34$\to$57.65 & 0.0476$\to$0.0587 & 93.61$\to$89.00 & 4154$\to$3702 \\
WeMM 9B (official) & 57.95$\to$57.37 & 0.0932$\to$0.1253 & 86.71$\to$86.48 & 3915$\to$3687 \\
Direct InfoNCE & 86.74$\to$86.20 & 0.7919$\to$0.7780 & 90.16$\to$95.12 & 2006$\to$2332 \\
Direct Within & 91.85$\to$89.21 & 0.8805$\to$0.8312 & 95.57$\to$97.14 & 2092$\to$2387 \\
Direct Cross & 93.86$\to$89.71 & 0.9081$\to$0.8408 & 97.80$\to$97.94 & 2174$\to$2453 \\
Stage1 InfoNCE & 56.62$\to$56.14 & 0.0600$\to$0.0596 & 90.51$\to$88.60 & 3939$\to$3766 \\
No-instruction Hier & 57.86$\to$57.27 & 0.0382$\to$0.0403 & 94.12$\to$90.96 & 3953$\to$3682 \\
Within + Replay & 90.85$\to$88.49 & 0.8653$\to$0.8215 & 94.92$\to$96.82 & 2018$\to$2326 \\
Cross + Replay & 92.31$\to$88.97 & 0.8856$\to$0.8289 & 96.44$\to$97.49 & 2122$\to$2389 \\
\bottomrule
\end{tabular}}
\endgroup
\end{table}

\begin{table}[!htbp]
\centering
\caption{UMAP comparison: generic category clustering. Each cell reports full dimensions $\to$ UMAP.}
\label{tab:umap-generic}
\begingroup
\scriptsize
\setlength{\tabcolsep}{2pt}
\resizebox{\linewidth}{!}{%
\begin{tabular}{@{}lrrrr@{}}
\toprule
Model & $V$ & ARI & Coverage & Clusters \\
\midrule
Gemini Embedding 2 & 83.40$\to$75.27 & 0.4274$\to$0.2019 & 84.03$\to$90.12 & 70$\to$340 \\
Qwen3-VL 2B (official) & 86.08$\to$74.14 & 0.4797$\to$0.1686 & 85.71$\to$87.64 & 75$\to$334 \\
Qwen3-VL 8B (official) & 89.32$\to$74.83 & 0.6000$\to$0.1847 & 91.68$\to$88.83 & 52$\to$328 \\
WeMM 2B (official) & 87.24$\to$75.51 & 0.5305$\to$0.2132 & 87.46$\to$89.88 & 70$\to$325 \\
WeMM 9B (official) & 86.67$\to$73.78 & 0.4955$\to$0.1559 & 85.58$\to$88.11 & 75$\to$345 \\
Direct InfoNCE & 71.02$\to$72.71 & 0.1423$\to$0.1447 & 76.12$\to$93.37 & 276$\to$489 \\
Direct Within & 75.46$\to$73.05 & 0.2528$\to$0.1471 & 84.59$\to$94.38 & 263$\to$507 \\
Direct Cross & 85.02$\to$75.49 & 0.5141$\to$0.1958 & 88.85$\to$93.58 & 123$\to$382 \\
Stage1 InfoNCE & 95.88$\to$74.02 & 0.8831$\to$0.1652 & 97.39$\to$90.90 & 68$\to$444 \\
No-instruction Hier & 96.21$\to$76.33 & 0.8711$\to$0.2341 & 98.36$\to$91.62 & 59$\to$347 \\
Within + Replay & 96.96$\to$76.56 & 0.8878$\to$0.2425 & 98.68$\to$92.13 & 56$\to$355 \\
Cross + Replay & 97.08$\to$75.12 & 0.8879$\to$0.1886 & 98.99$\to$90.30 & 55$\to$356 \\
\bottomrule
\end{tabular}}
\endgroup
\end{table}

\paragraph{Recorded instruction and attribute-order comparisons.}

\begin{table}[!ht]
\centering
\caption{Prefix, instruction-arity, and no-instruction configuration comparisons.}
\label{tab:historical-instruction-controls}
\begingroup\scriptsize
\setlength{\tabcolsep}{2pt}
\begin{tabular*}{\linewidth}{@{\extracolsep{\fill}}l*{8}{r}@{}}
\toprule
Model & 64 & 128 & Full & 1D & 2D & 3D & No-inst. & $\Delta V$ \\
\midrule
InfoNCE MRL frozen (old) & 93.39 & 93.46 & 93.61 & 92.65 & 94.65 & 94.37 & 41.93 & 51.68 \\
InfoNCE MRL unfrozen (old) & 93.04 & 93.22 & 93.25 & 91.92 & 94.62 & 94.86 & 42.05 & 51.20 \\
Direct InfoNCE full & 93.08 & 93.57 & 93.24 & 91.86 & 94.64 & 95.00 & 42.35 & 50.89 \\
Within MRL frozen (old) & 94.94 & 94.72 & 94.83 & 93.43 & 96.26 & 96.46 & 46.27 & 48.56 \\
Within MRL unfrozen (old) & 94.77 & 94.67 & 94.81 & 93.18 & 96.44 & 96.92 & 43.75 & 51.05 \\
Within system MRL (old) & 94.64 & 94.75 & 94.52 & 92.64 & 96.57 & 96.10 & 43.90 & 50.63 \\
Within schema-anchor (old) & 94.74 & 94.79 & 94.83 & 92.96 & 96.64 & 97.82 & 48.89 & 45.94 \\
Within unit-mean (old) & 94.74 & 94.56 & 94.70 & 92.72 & 96.70 & 97.29 & 47.51 & 47.19 \\
Tongyi Flash (recorded configuration) & 43.16 & 44.02 & 45.62 & 39.13 & 51.71 & 57.01 & 45.60 & 0.02 \\
Tongyi Plus (recorded configuration) & 46.64 & 47.55 & 47.45 & 40.68 & 53.61 & 60.54 & 46.75 & 0.71 \\
Gemini Embedding 2 & 45.63 & 48.25 & 49.68 & 43.59 & 55.14 & 61.95 & 49.00 & 0.68 \\
Direct Within full & 94.86 & 94.75 & 95.01 & 93.16 & 96.97 & 96.83 & 43.16 & 51.85 \\
Cross MRL frozen (old) & 94.78 & 94.81 & 95.03 & 93.42 & 96.63 & 97.28 & 44.31 & 50.72 \\
\bottomrule
\end{tabular*}
\endgroup
\par\smallskip
\begin{minipage}{\linewidth}\scriptsize
All entries are V-measure on a 0--100 scale; $\Delta V$ is in percentage points. 64/128 are prefix truncations, not necessarily MRL-trained representations. 1D/2D/3D use full-dimensional instructed embeddings over 230/189/30 tasks; Full and No-inst. average all 449 tasks. Per-model parameter selection uses evaluation labels; instructed and uninstructed configurations share the selected parameters. Their recorded missing-data handling does not establish a strictly matched fixed-pool ablation or cross-instruction relation recovery. Separate checkpoints are explicitly labeled. Flash/Plus input-fusion equivalence is not established by the available records. Different arities use different pools.
\end{minipage}
\end{table}

\paragraph{Batch composition.}
\label{sec:core-ablation}
The following three tables report nine completed CM24 configurations using the task-wise, pooled, and generic metric definitions from the main text. These comparisons concern overall training configurations; loss and sample-organization effects are not independently isolated.

\begin{table}[!htbp]
\centering
\caption{CM24 batch composition: task-wise conditional clustering.}
\label{tab:cm24-results}
\begingroup
\scriptsize
\setlength{\tabcolsep}{2pt}
\renewcommand{\arraystretch}{1.10}
\begin{tabular*}{\linewidth}{@{\extracolsep{\fill}}l*{6}{r}@{}}
\toprule
Method & $V\uparrow$ & ARI$\uparrow$ & $Q\uparrow$ & \shortstack{Coverage\\$\uparrow$} & \shortstack{Mean clusters\\$\to4.73$} & $R_{\mathrm{dist}}\downarrow$ \\
\midrule
$C2M12$ & 90.12 & 0.8333 & 0.5719 & 97.50 & 4.59 & 0.1396 \\
$C3M8$ & 92.72 & 0.8817 & 0.5443 & 97.53 & 4.74 & 0.1344 \\
$C4M6$ & 92.77 & 0.8836 & 0.5474 & 97.54 & 4.77 & 0.1296 \\
$C6M4$ & 92.55 & 0.8831 & 0.5416 & 97.14 & 4.74 & 0.1323 \\
Mixed C/M & 93.17 & 0.8922 & 0.5420 & 97.83 & 4.86 & \textbf{0.1294} \\
\midrule
Multi-Unit balanced (Within) & 92.82 & 0.8880 & 0.5443 & 97.54 & 4.82 & 0.1325 \\
Multi-Unit balanced (Cross) & 92.79 & 0.8882 & 0.5396 & 97.41 & 4.80 & 0.1341 \\
Single/Multi 5:5 (Within) & 92.83 & 0.8868 & 0.5459 & 97.35 & 4.78 & 0.1319 \\
Single/Multi 5:5 (Cross) & 93.10 & 0.8914 & 0.5439 & 98.25 & 4.86 & 0.1352 \\
\bottomrule
\end{tabular*}
\endgroup
\par\smallskip
\begin{minipage}{\linewidth}
\scriptsize
Metric definitions follow Appendix~\ref{app:evaluation-details}. Embeddings use full dimensionality (2,048 for Qwen/WeMM 2B and 4,096 for the other models); V and coverage are percentages. Noise $-1$ is retained for scoring but excluded from cluster counts. Direct starts from the official backbone; Within/Cross denote the comparison scope. Earlier replay checkpoints are separate configurations from Stage 3. Evaluation labels participate in clustering-parameter selection; -- denotes an unreported result. Scores and cluster counts are averaged equally over 449 tasks.
For this appendix batch, evaluation ground-truth labels were used in HDBSCAN parameter selection; parameters have not been frozen using an independent validation set. These are therefore not fixed-parameter held-out test results.
\end{minipage}
\end{table}

\begin{table}[!htbp]
\centering
\caption{CM24 batch composition: pooled conditional clustering.}
\label{tab:cm24-results-pooled}
\begingroup
\scriptsize
\setlength{\tabcolsep}{2pt}
\renewcommand{\arraystretch}{1.10}
\begin{tabular*}{\linewidth}{@{\extracolsep{\fill}}l*{7}{r}@{}}
\toprule
Method & $V\uparrow$ & $Q\uparrow$ & \shortstack{Coverage\\$\uparrow$} & \shortstack{Clusters\\$\to2123$} & $R_{\mathrm{dist}}\downarrow$ & \shortstack{Align-\\ment$\downarrow$} & \shortstack{Unifor-\\mity$\downarrow$} \\
\midrule
$C2M12$ & 86.80 & 0.3255 & 93.24 & 1887 & 0.5478 & 0.0525 & -2.1646 \\
$C3M8$ & 88.70 & 0.3636 & 93.07 & 1962 & 0.4253 & 0.0815 & -2.6152 \\
$C4M6$ & 89.20 & 0.3455 & 92.72 & 1935 & 0.4396 & 0.0797 & -2.6548 \\
$C6M4$ & 89.51 & 0.3579 & 93.49 & 1965 & 0.4123 & 0.0944 & -2.9466 \\
Mixed C/M & 89.48 & 0.3592 & 94.39 & 2018 & 0.4148 & 0.0800 & -2.6738 \\
\midrule
Multi-Unit balanced (Within) & 89.72 & 0.3621 & 93.60 & 1973 & 0.4141 & 0.0904 & -2.8390 \\
Multi-Unit balanced (Cross) & 91.55 & 0.4989 & 96.42 & 2083 & \textbf{0.2497} & 0.0875 & -3.5696 \\
Single/Multi 5:5 (Within) & 90.02 & 0.3623 & 93.83 & 1938 & 0.4189 & 0.0870 & -2.7677 \\
Single/Multi 5:5 (Cross) & 92.22 & 0.4970 & 96.98 & 2127 & 0.2652 & 0.0872 & -3.4955 \\
\bottomrule
\end{tabular*}
\endgroup
\par\smallskip
\begin{minipage}{\linewidth}
\scriptsize
Metric definitions follow Appendix~\ref{app:evaluation-details}. All configurations use full 4,096-dimensional embeddings. V and coverage are percentages. Noise label $-1$ is retained for scoring but excluded from cluster counts. Within/Cross denote the contrastive comparison scope. The pool contains 14,997 instructed records and 2,123 task-specific reference groups. V and coverage are task-macro averages computed after joint clustering, while Clusters is the global number of non-noise predicted clusters. HDBSCAN parameters are selected using evaluation labels; the reported results follow this label-assisted selection protocol.
\end{minipage}
\end{table}

\begin{table}[!htbp]
\centering
\caption{CM24 batch composition: generic category clustering.}
\label{tab:cm24-results-generic}
\begingroup
\scriptsize
\setlength{\tabcolsep}{2pt}
\renewcommand{\arraystretch}{1.10}
\begin{tabular*}{\linewidth}{@{\extracolsep{\fill}}l*{4}{r}@{}}
\toprule
Method & $V\uparrow$ & $Q\uparrow$ & Coverage$\uparrow$ & Clusters$\to50$ \\
\midrule
$C2M12$ & 63.25 & 0.0621 & 70.52 & 337 \\
$C3M8$ & 61.95 & 0.0458 & 72.64 & 566 \\
$C4M6$ & 60.15 & 0.0459 & 68.97 & 552 \\
$C6M4$ & 59.57 & 0.0458 & 73.07 & 1091 \\
Mixed C/M & 60.39 & 0.0465 & 69.72 & 581 \\
\midrule
Multi-Unit balanced (Within) & 61.24 & 0.0503 & 71.21 & 601 \\
Multi-Unit balanced (Cross) & 72.71 & 0.1188 & 76.47 & 201 \\
Single/Multi 5:5 (Within) & 60.19 & 0.0452 & 71.30 & 603 \\
Single/Multi 5:5 (Cross) & 71.90 & 0.1156 & 81.35 & 296 \\
\bottomrule
\end{tabular*}
\endgroup
\par\smallskip
\begin{minipage}{\linewidth}
\scriptsize
Metric definitions follow Appendix~\ref{app:evaluation-details}. Embeddings use full dimensionality (2,048 for Qwen/WeMM 2B and 4,096 for the other models); V and coverage are percentages. Noise $-1$ is retained for scoring but excluded from cluster counts. Direct starts from the official backbone; Within/Cross denote the comparison scope. Earlier replay checkpoints are separate configurations from Stage 3. Evaluation labels participate in clustering-parameter selection; -- denotes an unreported result. Category evaluation uses 5,373 distinct products with 50 leaf-category labels; Clusters is a global count.
For this appendix batch, evaluation ground-truth labels were used in HDBSCAN parameter selection; parameters have not been frozen using an independent validation set. These are therefore not fixed-parameter held-out test results.
\end{minipage}
\end{table}

\FloatBarrier

\subsection{Additional Backbones}

The advantage of structured multi-positive training extends across backbones and model sizes. On ProductMIC, Direct Within and Direct Cross outperform Direct InfoNCE in task-wise V-measure for Qwen 2B, WeMM 2B, and WeMM 9B (Table \ref{tab:supplement-model-results}). The corresponding scores are 93.17/93.54 versus 90.02 for Qwen 2B, 93.04/93.59 versus 89.69 for WeMM 2B, and 94.79/94.76 versus 91.96 for WeMM 9B. Direct Cross also achieves higher pooled V-measure and generic category V-measure than Direct Within on all three backbones (Tables \ref{tab:supplement-model-results-pooled} and \ref{tab:supplement-model-results-generic}), extending the local--global pattern observed on Qwen 8B.

\begin{table}[!htbp]
\centering
\caption{Additional model evaluation: task-wise conditional clustering.}
\label{tab:supplement-model-results}
\begingroup
\scriptsize
\setlength{\tabcolsep}{2pt}
\renewcommand{\arraystretch}{1.10}
\begin{tabular*}{\linewidth}{@{\extracolsep{\fill}}l*{6}{r}@{}}
\toprule
Method & $V\uparrow$ & ARI$\uparrow$ & $Q\uparrow$ & \shortstack{Coverage\\$\uparrow$} & \shortstack{Mean clusters\\$\to4.73$} & $R_{\mathrm{dist}}\downarrow$ \\
\midrule
Qwen 2B Cross+Replay & 92.76 & 0.8842 & 0.4974 & 97.90 & 4.83 & 0.1765 \\
Qwen 2B Hier & 41.21 & 0.1416 & 0.1519 & 78.79 & 6.65 & 0.9948 \\
WeMM 2B Hier & 42.60 & 0.1468 & 0.1183 & 76.95 & 6.99 & 0.9678 \\
WeMM 9B Hier & 44.01 & 0.1619 & 0.1238 & 79.60 & 7.13 & 0.9500 \\
Earlier InfoNCE replay & 93.22 & 0.8939 & 0.4689 & 98.04 & 4.92 & 0.2208 \\
Hier SupCon V3 (earlier checkpoint) & 44.34 & 0.1750 & 0.1631 & 82.54 & 6.84 & 0.9627 \\
Earlier Cross replay & 94.09 & 0.9102 & 0.5474 & 98.00 & 4.85 & 0.1186 \\
Stage1 InfoNCE V3 (earlier checkpoint) & 42.90 & 0.1684 & 0.1800 & 81.78 & 6.35 & 0.9887 \\
\midrule
Stage2 InfoNCE (4096) & 47.65 & 0.1950 & 0.1063 & 80.90 & 7.52 & 0.9027 \\
Stage3 InfoNCE+Replay (4096) & 93.27 & 0.8933 & 0.4132 & 97.28 & 4.85 & 0.2904 \\
WeMM 2B Cross+Replay & 91.87 & 0.8713 & 0.4909 & 97.98 & 4.76 & 0.1813 \\
WeMM 9B Cross+Replay TP2 & 94.70 & 0.9155 & 0.5464 & 98.70 & 4.84 & 0.1127 \\
\midrule
\multicolumn{7}{@{}l}{\textbf{Direct conditional training}} \\
Qwen 2B Direct InfoNCE & 90.02 & 0.8351 & 0.3725 & 97.01 & 4.84 & 0.3282 \\
Qwen 2B Direct Within & 93.17 & 0.8941 & 0.5094 & 98.01 & 4.85 & 0.1567 \\
Qwen 2B Direct Cross & 93.54 & 0.8995 & 0.5129 & 98.05 & 4.86 & 0.1545 \\
WeMM 2B Direct InfoNCE & 89.69 & 0.8357 & 0.3517 & 96.63 & 4.90 & 0.3430 \\
WeMM 2B Direct Within & 93.04 & 0.8892 & 0.5049 & 98.09 & 4.82 & 0.1588 \\
WeMM 2B Direct Cross & 93.59 & 0.8964 & 0.5090 & 98.58 & 4.82 & 0.1600 \\
WeMM 9B Direct InfoNCE & 91.96 & 0.8733 & 0.4367 & 97.58 & 5.00 & 0.2530 \\
WeMM 9B Direct Within & 94.79 & 0.9184 & 0.5530 & 98.58 & 4.84 & \textbf{0.1042} \\
WeMM 9B Direct Cross & 94.76 & 0.9194 & 0.5552 & 98.48 & 4.85 & 0.1063\\
\bottomrule
\end{tabular*}
\endgroup
\par\smallskip
\begin{minipage}{\linewidth}
\scriptsize
Metric definitions follow Appendix~\ref{app:evaluation-details}. Embeddings use full dimensionality (2,048 for Qwen/WeMM 2B and 4,096 for the other models); V and coverage are percentages. Noise $-1$ is retained for scoring but excluded from cluster counts. Direct starts from the official backbone; Within/Cross denote the comparison scope. Earlier replay checkpoints are separate configurations from Stage 3. Evaluation labels participate in clustering-parameter selection; -- denotes an unreported result. Scores and cluster counts are averaged equally over 449 tasks.
For this appendix batch, evaluation ground-truth labels were used in HDBSCAN parameter selection; parameters have not been frozen using an independent validation set. These are therefore not fixed-parameter held-out test results.
\end{minipage}
\end{table}

\begin{table}[!htbp]
\centering
\caption{Additional model evaluation: pooled conditional clustering.}
\label{tab:supplement-model-results-pooled}
\begingroup
\scriptsize
\setlength{\tabcolsep}{2pt}
\renewcommand{\arraystretch}{1.10}
\begin{tabular*}{\linewidth}{@{\extracolsep{\fill}}l*{7}{r}@{}}
\toprule
Method & $V\uparrow$ & $Q\uparrow$ & \shortstack{Coverage\\$\uparrow$} & \shortstack{Clusters\\$\to2123$} & $R_{\mathrm{dist}}\downarrow$ & \shortstack{Align-\\ment$\downarrow$} & \shortstack{Unifor-\\mity$\downarrow$} \\
\midrule
Qwen 2B Cross+Replay & 90.43 & 0.4182 & 95.80 & 2089 & 0.3390 & 0.0933 & -3.1460 \\
Qwen 2B Hier & 58.59 & 0.0817 & 94.39 & 3978 & 1.2767 & 0.0514 & -2.1335 \\
WeMM 2B Hier & 59.36 & 0.0842 & 94.99 & 4156 & 1.2166 & 0.0646 & -1.9737 \\
WeMM 9B Hier & 59.30 & 0.0894 & 95.32 & 4087 & 1.2115 & 0.0606 & -2.0496 \\
Earlier InfoNCE replay & 87.80 & 0.2766 & 92.78 & 2054 & 0.4680 & 0.1886 & -3.2836 \\
Hier SupCon V3 (earlier checkpoint) & 57.95 & 0.0835 & 94.64 & 3893 & 1.2596 & 0.0502 & -2.0962 \\
Earlier Cross replay & 92.57 & 0.4851 & 96.74 & 2116 & 0.2586 & 0.0660 & -3.0549 \\
Stage1 InfoNCE V3 (earlier checkpoint) & 56.23 & 0.0589 & 89.92 & 3962 & 1.2680 & 0.0501 & -3.0653 \\
\midrule
Stage2 InfoNCE (4096) & 58.74 & 0.0721 & 93.04 & 4093 & 1.1072 & 0.2198 & -3.0735 \\
Stage3 InfoNCE+Replay (4096) & 85.76 & 0.1899 & 88.99 & 1908 & 0.5371 & 0.2853 & -3.4670 \\
WeMM 2B Cross+Replay & 89.87 & 0.4091 & 95.88 & 2079 & 0.3425 & 0.0940 & -3.1049 \\
WeMM 9B Cross+Replay TP2 & 93.40 & 0.4953 & 97.52 & 2147 & \textbf{0.2377} & 0.0631 & -3.2366 \\
\midrule
\multicolumn{8}{@{}l}{\textbf{Direct conditional training}} \\
Qwen 2B Direct InfoNCE & 81.06 & 0.1457 & 86.13 & 1931 & 0.6090 & 0.2671 & -3.1257 \\
Qwen 2B Direct Within & 89.31 & 0.2993 & 94.02 & 1976 & 0.4394 & 0.0798 & -2.3514 \\
Qwen 2B Direct Cross & 92.30 & 0.4503 & 96.72 & 2125 & 0.2834 & 0.0811 & -3.1554 \\
WeMM 2B Direct InfoNCE & 78.43 & 0.1102 & 83.73 & 1923 & 0.6625 & 0.3571 & -3.5040 \\
WeMM 2B Direct Within & 90.10 & 0.3326 & 94.81 & 2053 & 0.3989 & 0.0803 & -2.4023 \\
WeMM 2B Direct Cross & 92.34 & 0.4386 & 96.79 & 2127 & 0.2899 & 0.0827 & -3.2020 \\
WeMM 9B Direct InfoNCE & 84.94 & 0.2137 & 90.13 & 2080 & 0.5078 & 0.2282 & -3.2537 \\
WeMM 9B Direct Within & 92.21 & 0.4279 & 95.79 & 2075 & 0.2955 & 0.0562 & -2.4793 \\
WeMM 9B Direct Cross & 93.75 & 0.5255 & 98.06 & 2159 & 0.1940 & 0.0580 & -3.2726 \\
\bottomrule
\end{tabular*}
\endgroup
\par\smallskip
\begin{minipage}{\linewidth}
\scriptsize
Metric definitions follow Appendix~\ref{app:evaluation-details}. Embeddings use full dimensionality (2,048 for Qwen/WeMM 2B and 4,096 for the other models); V and coverage are percentages. Noise $-1$ is retained for scoring but excluded from cluster counts. Direct starts from the official backbone; Within/Cross denote the comparison scope. Earlier replay checkpoints are separate configurations from Stage 3. Evaluation labels participate in clustering-parameter selection; -- denotes an unreported result. The pool contains 14,997 instructed records and 2,123 reference groups; V and coverage are task-macro averages, while Clusters is a global count. WeMM 9B Direct Cross reuses its task-wise-selected parameters without an additional pooled search.
For this appendix batch, evaluation ground-truth labels were used in HDBSCAN parameter selection; parameters have not been frozen using an independent validation set. These are therefore not fixed-parameter held-out test results.

\end{minipage}
\end{table}

\begin{table}[!htbp]
\centering
\caption{Additional model evaluation: generic category clustering.}
\label{tab:supplement-model-results-generic}
\begingroup
\scriptsize
\setlength{\tabcolsep}{2pt}
\renewcommand{\arraystretch}{1.10}
\begin{tabular*}{\linewidth}{@{\extracolsep{\fill}}l*{4}{r}@{}}
\toprule
Method & $V\uparrow$ & $Q\uparrow$ & Coverage$\uparrow$ & Clusters$\to50$ \\
\midrule
Qwen 2B Cross+Replay & 97.67 & 0.6963 & 99.03 & 53 \\
Qwen 2B Hier & 96.46 & 0.7054 & 98.34 & 58 \\
WeMM 2B Hier & 96.49 & 0.6493 & 97.75 & 51 \\
WeMM 9B Hier & 97.40 & 0.7427 & 99.74 & 50 \\
Earlier InfoNCE replay & 94.49 & 0.4669 & 97.54 & 63 \\
Hier SupCon V3 (earlier checkpoint) & 96.49 & 0.7880 & 98.88 & 57 \\
Earlier Cross replay & 97.35 & 0.7425 & 99.76 & 50 \\
Stage1 InfoNCE V3 (earlier checkpoint) & 96.21 & 0.8474 & 97.97 & 62 \\
\midrule
Stage2 InfoNCE (4096) & 94.87 & 0.5823 & 97.58 & 55 \\
Stage3 InfoNCE+Replay (4096) & 94.87 & 0.3224 & 98.73 & 56 \\
WeMM 2B Cross+Replay & 95.58 & 0.4711 & 98.94 & 50 \\
WeMM 9B Cross+Replay TP2 & 97.79 & 0.6974 & 99.81 & 50 \\
\midrule
\multicolumn{5}{@{}l}{\textbf{Direct conditional training}} \\
Qwen 2B Direct InfoNCE & 63.73 & 0.0201 & 68.14 & 320 \\
Qwen 2B Direct Within & 68.57 & 0.0455 & 76.94 & 300 \\
Qwen 2B Direct Cross & 80.83 & 0.1270 & 85.99 & 120 \\
WeMM 2B Direct InfoNCE & 70.04 & 0.0279 & 72.96 & 227 \\
WeMM 2B Direct Within & 74.20 & 0.0613 & 82.10 & 256 \\
WeMM 2B Direct Cross & 84.04 & 0.1423 & 89.91 & 108 \\
WeMM 9B Direct InfoNCE & 70.52 & 0.0302 & 74.15 & 224 \\
WeMM 9B Direct Within & 69.20 & 0.0536 & 78.52 & 322 \\
WeMM 9B Direct Cross & 82.77 & 0.1190 & 85.99 & 126 \\
\bottomrule
\end{tabular*}
\endgroup
\par\smallskip
\begin{minipage}{\linewidth}
\scriptsize
Metric definitions follow Appendix~\ref{app:evaluation-details}. Embeddings use full dimensionality (2,048 for Qwen/WeMM 2B and 4,096 for the other models); V and coverage are percentages. Noise $-1$ is retained for scoring but excluded from cluster counts. Direct starts from the official backbone; Within/Cross denote the comparison scope. Earlier replay checkpoints are separate configurations from Stage 3. Evaluation labels participate in clustering-parameter selection; -- denotes an unreported result. Category evaluation uses 5,373 distinct products with 50 leaf-category labels; Clusters is a global count.
For this appendix batch, evaluation ground-truth labels were used in HDBSCAN parameter selection; parameters have not been frozen using an independent validation set. These are therefore not fixed-parameter held-out test results.

\end{minipage}
\end{table}

\clearpage
\subsection{Full Pooled and Category Results}
\label{app:full-scopes}

\begin{table}[!htbp]
\centering
\caption{Pooled conditional clustering with task-specific ground-truth groups $(U,y_i^U)$ retained across $449$ tasks ($2{,}123$ groups). Model groups follow Table~\ref{tab:embedding-main}.}
\label{tab:embedding-pooled}
\begingroup
\scriptsize
\setlength{\tabcolsep}{2pt}
\renewcommand{\arraystretch}{1.10}
\begin{tabular*}{\linewidth}{@{\extracolsep{\fill}}l*{6}{r}@{}}
\toprule
Method & $V\uparrow$ & ARI$\uparrow$ & $Q\uparrow$ & Coverage$\uparrow$ & Clusters & $R_{\mathrm{dist}}\downarrow$ \\
\midrule
\multicolumn{7}{@{}l}{\textbf{General-purpose models}} \\
Tongyi Flash & 59.45 & 0.0446 & 0.0802 & 95.01 & 3999 & 1.1008 \\
Tongyi Plus & 59.32 & 0.0505 & 0.0677 & 94.13 & 3979 & 1.0788 \\
Gemini Embedding 2 & 59.04 & 0.0417 & 0.0504 & 94.48 & 3806 & 1.0030 \\
Qwen3-VL 2B (official) & 59.37 & 0.0354 & 0.0669 & 94.60 & 4164 & 1.0773 \\
Qwen3-VL 8B (official) & 59.21 & 0.0631 & 0.0364 & 91.59 & 4002 & 0.9705 \\
WeMM 2B (official) & 59.34 & 0.0476 & 0.0443 & 93.61 & 4154 & 0.9731 \\
WeMM 9B (official) & 57.95 & 0.0932 & 0.0213 & 86.71 & 3915 & 0.9147 \\
\midrule
\multicolumn{7}{@{}l}{\textbf{Direct conditional training (Qwen3-VL 8B)}} \\
InfoNCE & 86.74 & 0.7919 & 0.2266 & 90.16 & 2006 & 0.4842 \\
Within SupCon & 91.85 & 0.8805 & 0.4279 & 95.57 & 2092 & 0.2921 \\
Cross SupCon & \textbf{93.86} & \textbf{0.9081} & \textbf{0.5207} & \textbf{97.80} & 2174 & \textbf{0.1940} \\
\midrule
\multicolumn{7}{@{}l}{\textbf{Staged training (InfoNCE)}} \\
Stage 1 (generic) & 56.62 & 0.0600 & 0.0622 & 90.51 & 3939 & 1.2567 \\
Stage 2 (generic) & 58.74 & 0.0474 & 0.0721 & 93.04 & 4093 & 1.1072 \\
Stage 3 (instruction + replay) & 85.76 & 0.7749 & 0.1899 & 88.99 & 1908 & 0.5371 \\
\midrule
\multicolumn{7}{@{}l}{\textbf{Staged training (SupCon)}} \\
Hierarchical SupCon (generic) & 57.86 & 0.0382 & 0.0785 & 94.12 & 3953 & 1.2733 \\
Within + generic replay & 90.85 & 0.8653 & 0.4021 & 94.92 & 2018 & 0.3380 \\
Cross + generic replay & 92.31 & 0.8856 & 0.4938 & 96.44 & \textbf{2122} & 0.2513 \\
\bottomrule
\end{tabular*}
\endgroup
\par\smallskip
\begin{minipage}{\linewidth}
\scriptsize
Metric definitions and dimensionality settings follow Appendix~\ref{app:evaluation-details}.
InfoNCE Stages 2 and 3 use non-MRL checkpoints: Stage 2 uses generic training; Stage 3 uses conditional training with generic replay and differs from the earlier InfoNCE replay checkpoint.
Alignment and Uniformity appear in Appendix Table~\ref{tab:embedding-pooled-geometry}.
\end{minipage}
\end{table}

\begin{table}[!htbp]
\centering
\caption{Generic category clustering with $50$ leaf categories as targets. Model groups follow Table~\ref{tab:embedding-main}; generic inputs specify no target attribute.}
\label{tab:embedding-generic}
\begingroup
\scriptsize
\setlength{\tabcolsep}{2pt}
\renewcommand{\arraystretch}{1.10}
\begin{tabular*}{\linewidth}{@{\extracolsep{\fill}}l*{6}{r}@{}}
\toprule
Method & $V\uparrow$ & ARI$\uparrow$ & $Q\uparrow$ & Coverage$\uparrow$ & Clusters & $R_{\mathrm{dist}}\downarrow$ \\
\midrule
\multicolumn{7}{@{}l}{\textbf{General-purpose models}} \\
Tongyi Flash & 84.85 & 0.4199 & 0.1214 & 82.93 & 79 & 0.6217 \\
Tongyi Plus & 87.99 & 0.5559 & 0.1393 & 88.57 & 74 & 0.6002 \\
Gemini Embedding 2 & 83.40 & 0.4274 & 0.0609 & 84.03 & 70 & 0.7234 \\
Qwen3-VL 2B (official) & 86.08 & 0.4797 & 0.1264 & 85.71 & 75 & 0.6452 \\
Qwen3-VL 8B (official) & 89.32 & 0.6000 & 0.1132 & 91.68 & \textbf{52} & 0.6180 \\
WeMM 2B (official) & 87.24 & 0.5305 & 0.1276 & 87.46 & 70 & 0.6578 \\
WeMM 9B (official) & 86.67 & 0.4955 & 0.1050 & 85.58 & 75 & 0.6858 \\
\midrule
\multicolumn{7}{@{}l}{\textbf{Direct conditional training (Qwen3-VL 8B)}} \\
InfoNCE & 71.02 & 0.1423 & 0.0297 & 76.12 & 276 & 0.7621 \\
Within SupCon & 75.46 & 0.2528 & 0.0618 & 84.59 & 263 & 0.7224 \\
Cross SupCon & 85.02 & 0.5141 & 0.1704 & 88.85 & 123 & 0.5698 \\
\midrule
\multicolumn{7}{@{}l}{\textbf{Staged training (InfoNCE)}} \\
Stage 1 (generic) & 95.88 & 0.8831 & \textbf{0.8433} & 97.39 & 68 & \textbf{0.0921} \\
Stage 2 (generic) & 94.87 & 0.7734 & 0.5823 & 97.58 & 55 & 0.2894 \\
Stage 3 (instruction + replay) & 94.87 & 0.7732 & 0.3224 & 98.73 & 56 & 0.4938 \\
\midrule
\multicolumn{7}{@{}l}{\textbf{Staged training (SupCon)}} \\
Hierarchical SupCon (generic) & 96.21 & 0.8711 & 0.7973 & 98.36 & 59 & 0.1166 \\
Within + generic replay & 96.96 & 0.8878 & 0.7430 & 98.68 & 56 & 0.1733 \\
Cross + generic replay & \textbf{97.08} & \textbf{0.8879} & 0.7457 & \textbf{98.99} & 55 & 0.1531 \\
\bottomrule
\end{tabular*}
\endgroup
\par\smallskip
\begin{minipage}{\linewidth}
\scriptsize
Metric definitions and dimensionality settings follow Appendix~\ref{app:evaluation-details}.
InfoNCE Stages 2 and 3 use non-MRL checkpoints: Stage 2 uses generic training; Stage 3 uses conditional training with generic replay and differs from the earlier InfoNCE replay checkpoint.
Alignment and Uniformity appear in Appendix Table~\ref{tab:embedding-generic-geometry}.
\end{minipage}
\end{table}

\subsection{Representation Diagnostics}
\label{app:main-geometry}

\begin{table}[htbp]
\centering\scriptsize
\caption{Pooled representation diagnostics. Model order, selection, and aggregation follow Table~\ref{tab:embedding-pooled}.}
\label{tab:embedding-pooled-geometry}
\begin{tabular*}{\linewidth}{@{\extracolsep{\fill}}lrr@{}}
\toprule
Method & \shortstack{Align-\\ment$\downarrow$} & \shortstack{Unifor-\\mity$\downarrow$} \\
\midrule
\multicolumn{3}{@{}l}{\textbf{General-purpose models}} \\
Tongyi Flash & 0.2533 & -1.3519 \\
Tongyi Plus & 0.2455 & -1.4391 \\
Gemini Embedding 2 & 0.2525 & -1.3503 \\
Qwen3-VL 2B (official) & 0.5255 & -2.7918 \\
Qwen3-VL 8B (official) & 0.5460 & -2.9168 \\
WeMM 2B (official) & 0.5162 & -2.5893 \\
WeMM 9B (official) & 0.4369 & -2.3597 \\
\midrule
\multicolumn{3}{@{}l}{\textbf{Direct conditional training (Qwen3-VL 8B)}} \\
InfoNCE & 0.2237 & -3.2742 \\
Within SupCon & 0.0529 & -2.5084 \\
Cross SupCon & 0.0590 & -3.2891 \\
\midrule
\multicolumn{3}{@{}l}{\textbf{Staged training (InfoNCE)}} \\
Stage 1 (generic) & 0.0530 & -3.0599 \\
Stage 2 (generic) & 0.2198 & -3.0735 \\
Stage 3 (instruction + replay) & 0.2853 & \textbf{-3.4670} \\
\midrule
\multicolumn{3}{@{}l}{\textbf{Staged training (SupCon)}} \\
Hierarchical SupCon (generic) & \textbf{0.0401} & -2.0250 \\
Within + generic replay & 0.0598 & -2.5471 \\
Cross + generic replay & 0.0641 & -3.1986 \\
\bottomrule
\end{tabular*}
\end{table}

\begin{table}[htbp]
\centering\scriptsize
\caption{Category representation diagnostics. Model order, selection, and aggregation follow Table~\ref{tab:embedding-generic}.}
\label{tab:embedding-generic-geometry}
\begin{tabular*}{\linewidth}{@{\extracolsep{\fill}}lrr@{}}
\toprule
Method & \shortstack{Align-\\ment$\downarrow$} & \shortstack{Unifor-\\mity$\downarrow$} \\
\midrule
\multicolumn{3}{@{}l}{\textbf{General-purpose models}} \\
Tongyi Flash & 0.3228 & -1.3505 \\
Tongyi Plus & 0.3336 & -1.4376 \\
Gemini Embedding 2 & 0.4251 & -1.4288 \\
Qwen3-VL 2B (official) & 0.8524 & -3.2855 \\
Qwen3-VL 8B (official) & 0.6874 & -2.8275 \\
WeMM 2B (official) & 0.7062 & -2.6859 \\
WeMM 9B (official) & 0.6659 & -2.3822 \\
\midrule
\multicolumn{3}{@{}l}{\textbf{Direct conditional training (Qwen3-VL 8B)}} \\
InfoNCE & 0.9990 & -2.9523 \\
Within SupCon & 0.4362 & -1.5023 \\
Cross SupCon & 0.6340 & -2.8466 \\
\midrule
\multicolumn{3}{@{}l}{\textbf{Staged training (InfoNCE)}} \\
Stage 1 (generic) & 0.1060 & -3.1969 \\
Stage 2 (generic) & 0.3525 & -3.1933 \\
Stage 3 (instruction + replay) & 0.7466 & \textbf{-3.5386} \\
\midrule
\multicolumn{3}{@{}l}{\textbf{Staged training (SupCon)}} \\
Hierarchical SupCon (generic) & \textbf{0.0652} & -2.0898 \\
Within + generic replay & 0.1245 & -2.4099 \\
Cross + generic replay & 0.1163 & -2.5801 \\
\bottomrule
\end{tabular*}
\end{table}
\clearpage

\clearpage

\paragraph{Postprocessing does not close the representation gap.}
Under the tested five-dimensional UMAP configuration, WeMM 9B improves from V=53.06 to 60.53, but remains well below Direct Within at 92.97 (95.01 without projection). Moreover, pooled V decreases for every reported configuration; Cross with replay produces 356 generic clusters after projection, compared with 55 before projection and 50 reference categories. Thus, this projection is not a reliable substitute for conditional representation learning and can fragment category structure. UMAP is label-free, but HDBSCAN parameters are reused rather than retuned, limiting this finding to the tested pipeline (Appendix Tables~\ref{tab:umap-task}--\ref{tab:umap-generic}).
\FloatBarrier

\section{External Conditional Methods}
\label{app:external-methods}
\label{app:supporting-analyses}
\subsection{External Implementations and Comparability}

External comparisons cover ABC, a clustering adaptation of CLAY, GCPM, Multi-Sub, and ITGC~\citep{schneider2025abc,lim2026clay,zhong2026guidelineclustering}. ABC and CLAY evaluate complementary routes to conditional representations. GCPM uses proxy text (G1), image--text with proxies (G2), and their extensions with MST--LLM semantic refinement (G3/G4), separating representation generation from subsequent cluster refinement.

All representation statistics use full native dimensions: 4,096 for ABC, CLAY, and GCPM; 80 concept coordinates for ITGC; and 8,192 for Multi-Sub. Predicted cluster counts exclude noise. $Q_{\mathrm{target}}$ uses $k=5,T=1$ on L2-normalized cosine distances; distance statistics are averaged over reference classes and then tasks. G1/G3 and G2/G4 share identical representations and therefore identical geometric metrics. Their $V$, ARI, coverage, and cluster counts describe initial HDBSCAN predictions for G1/G2 and refined predictions for G3/G4, respectively.

\paragraph{External-method results.}
All eight external configurations cover 449 task units and 14,997 object--instruction records. Table~\ref{tab:external-dimensions} reports overall performance and the breakdown by instruction dimensionality. GCPM G3 achieves the highest overall V-measure among these external configurations (74.85), followed by G4 (74.17). Among ABC, CLAY, ITGC, and Multi-Sub, ABC achieves the highest overall score (54.52), while CLAY exceeds ABC on three-attribute tasks (69.54 versus 65.86). Overall metrics are macro-averaged over the 449 tasks, with noise label $-1$ retained for scoring and coverage reported separately. The dimensional breakdown describes performance on the respective task pools, which differ across 1D, 2D, and 3D conditions.

The original G1/G2 configurations use HDBSCAN with minimum cluster size 2 and minimum samples 1 on L2-normalized embeddings. G1 and G2 achieve V-measures of 57.32 and 57.61; semantic refinement increases these by 17.54 and 16.56 percentage points, respectively. Refinement includes assigning initially unclustered objects, increasing the non-noise assignment fraction from 76.06/78.05 to 100 percent. The four-way comparison therefore measures the contribution of the complete refinement stage, including both cluster merging and noise assignment. G3/G4 coverage is marked as complete assignment rather than ranked against density-based rejection.

In the original benchmark configuration, Direct Within SupCon achieves V=95.01, exceeding GCPM G3 by 20.16 percentage points (Table~\ref{tab:embedding-main}). The originally reported image-only adaptations of ABC, CLAY, ITGC, and Multi-Sub use fixed HDBSCAN parameters $(2,1)$, whereas our embedding runs use label-assisted parameter selection. We therefore supplement these results with the shared selection protocol below to assess the effect of equalizing the final clustering search budget.

\begin{table}[!htbp]
\centering
\small
\setlength{\tabcolsep}{3pt}
\caption{External methods by instruction dimensionality under their original evaluation configurations. V-measure and coverage are percentages; ARI uses its original scale. Struck-through G3/G4 coverage marks complete assignment and is excluded from coverage ranking; the largest coverage among the remaining methods is bold.}
\label{tab:external-dimensions}
\begin{tabular}{lrrrrrr}
\toprule
Method & Overall $V$ & Coverage & ARI & 1D $V$ & 2D $V$ & 3D $V$ \\
\midrule
GCPM G1 & 57.32 & 76.06 & 0.3123 & 52.05 & 62.07 & 67.74 \\
GCPM G2 & 57.61 & 78.05 & 0.3083 & 52.59 & 61.62 & 70.81 \\
GCPM G3 & 74.85 & \sout{100.00} & 0.5786 & 75.09 & 74.04 & 78.21 \\
GCPM G4 & 74.17 & \sout{100.00} & 0.5683 & 75.30 & 72.36 & 76.87 \\
ABC & 54.52 & \textbf{86.80} & 0.3221 & 48.38 & 60.20 & 65.86 \\
CLAY & 51.86 & 85.99 & 0.2415 & 43.72 & 58.96 & 69.54 \\
ITGC & 46.57 & 80.82 & 0.2103 & 38.41 & 53.67 & 64.44 \\
Multi-Sub & 30.21 & 75.42 & 0.0728 & 25.59 & 34.31 & 39.83 \\
\bottomrule
\end{tabular}
\end{table}

\paragraph{Unified HDBSCAN selection protocol.}
We re-evaluate 12 sets of saved representations on the same 449 tasks and 14,997 records. All methods retain their full native dimensions and use float64 L2 normalization followed by Euclidean HDBSCAN (\texttt{hdbscan} 0.8.44, EOM), without dimensionality reduction. Pairwise distances are computed once per task and reused across candidates. Record order is fixed to the released evaluation manifest. Noise label $-1$ is retained as one predicted cluster, and coverage measures the fraction of objects assigned to non-noise clusters.

We compare three settings. \emph{Fixed} uses $(\texttt{min\_cluster\_size},\texttt{min\_samples})=(2,1)$ for every method. \emph{Shared-grid selection} evaluates the same 12 candidates, combining $\texttt{min\_cluster\_size}\in\{2,3,4,5\}$ with $ \texttt{min\_samples}\in\{\mathrm{None},1,2\}$; \texttt{None} sets minimum samples to minimum cluster size. Each model selects one parameter pair for all tasks by mean all-object V-measure, breaking exact ties by higher mean coverage and then fixed grid order. Selection and scoring use the same task pool in this setting. This shared criterion differs from the original composite selection criterion, so the new results are reported separately from the original benchmark scores.

\emph{Leaf-grouped cross-validation} divides the 50 leaf categories into five fixed folds of ten categories each. Each fold selects parameters using tasks from the other 40 categories and evaluates them on the ten held-out categories. All tasks from a leaf remain in the same fold, and product IDs do not overlap between selection and evaluation partitions. The folds are determined by a fixed hash ordering of leaf IDs. Reported scores average each task's out-of-fold result equally over all 449 tasks. This is a retrospective, leaf-disjoint validation of clustering-parameter selection on the existing benchmark.

\begin{table}[!htbp]
\centering
\small
\setlength{\tabcolsep}{4pt}
\caption{Conditional clustering under a unified HDBSCAN selection protocol. Fixed and Grid report V-measure with fixed $(2,1)$ parameters and shared-grid selection on the full task pool, respectively. CV reports five-fold leaf-grouped cross-validation. All metrics are task-macro averages. V-measure and coverage are percentages; ARI is unscaled. Unless specified otherwise, Direct configurations use Qwen3-VL 8B.}
\label{tab:external-unified-hdbscan}
\begin{tabular}{lrrrrr}
\toprule
Method & Fixed $V$ & Grid $V$ & CV $V$ & CV ARI & CV Cov. \\
\midrule
Qwen 8B Base & 50.89 & 50.89 & 50.89 & 0.2185 & 75.00 \\
WeMM 9B Base & 53.06 & 53.06 & 53.06 & 0.2418 & 74.06 \\
\midrule
GCPM G1 & 57.32 & 57.32 & 57.32 & 0.3123 & 76.06 \\
GCPM G2 & 57.61 & 57.61 & 57.61 & 0.3083 & 78.05 \\
ABC & 54.52 & 54.52 & 54.52 & 0.3221 & 86.80 \\
CLAY & 51.86 & 51.86 & 51.86 & 0.2415 & 85.99 \\
ITGC & 46.57 & 46.57 & 46.57 & 0.2103 & 80.82 \\
Multi-Sub & 30.21 & 30.21 & 30.21 & 0.0728 & 75.42 \\
\midrule
Direct InfoNCE & 87.93 & 94.12 & 94.12 & 0.9019 & \textbf{98.68} \\
Direct Within & 90.22 & \textbf{95.22} & \textbf{95.15} & \textbf{0.9226} & 98.33 \\
Direct Cross & \textbf{90.75} & 94.81 & 94.68 & 0.9172 & 98.44 \\
WeMM 9B Direct Cross & 90.39 & 94.76 & 94.37 & 0.9152 & 98.06 \\
\bottomrule
\end{tabular}
\end{table}

\paragraph{Results under matched selection rules.}
The advantage of directly trained representations persists across all three settings (Table~\ref{tab:external-unified-hdbscan}). ABC, CLAY, ITGC, Multi-Sub, and GCPM G1/G2 select their original $(2,1)$ configuration both on the full pool and in every cross-validation fold, leaving their scores unchanged within this grid. Direct Within achieves V=90.22 under the same fixed configuration, 95.22 under shared-grid selection, and 95.15 under leaf-grouped cross-validation. Direct InfoNCE and Direct Cross also retain strong out-of-fold scores of 94.12 and 94.68. Providing the external representations with the same HDBSCAN search budget therefore preserves the observed performance advantage. The close agreement between full-pool and out-of-fold scores further supports its stability across the evaluated leaf categories.

This comparison standardizes final clustering and parameter selection while preserving each method's saved representation. Input modalities and training procedures follow the original runs: the four external adaptations use image--instruction inputs, while our direct embedding configurations use product images, text, and instructions. For ITGC, we re-evaluate terminal clustering of the representation produced by its original feedback search. The results thus establish robustness to the tested clustering-selection protocol while retaining the original systems' input and training differences.

\paragraph{Consistency of GCPM refinement.}
For both GCPM branches, the selected initial partitions and noise masks match the original semantic-refinement inputs on all 449 tasks. The existing G3/G4 outputs therefore remain applicable and were re-scored without additional language-model calls. G3 retains V=74.85 and ARI=0.5786, while G4 retains V=74.17 and ARI=0.5683. Both retain complete assignment. These scores describe the existing refinement pipelines, whereas Table~\ref{tab:external-unified-hdbscan} compares the embedding-based clustering stage.

\paragraph{Generative workload and accounting scope.}
The unified clustering comparison reuses existing embeddings and generation outputs, leaving the recorded generative workload unchanged. Table~\ref{tab:external-llm-cost} summarizes the additional strong-LLM requests and tokens consumed by the evaluated pipelines.

One request denotes one independent generation prompt, regardless of batching. Counts include cached generation consumed by evaluation and exclude training annotation and unrelated exploration. CLAY generates conditional descriptions/concepts, Multi-Sub generates reference concepts, and ITGC uses three rounds of concept generation and feedback search. GCPM combines object-conditioned proxy generation with sequential cluster-merge
decisions. Proxy text-token counts are reconstructed from saved generation text and exclude image tokens; merge tokens are recorded at runtime. The reported proxy counts therefore measure text-token volume rather than complete multimodal input-token usage. G3 and G4 are alternative branches whose costs are not summed for a single deployment.

Request counts and token volumes characterize the generative workload. Multi-Sub additionally performs per-unit optimization, and ITGC performs iterative search. The available accounting does not provide comparable end-to-end timing or peak-memory measurements across methods. GCPM proxy-generation costs are shared by G1/G2, and zero additional generative requests does not imply zero embedding inference cost.

\begin{table}[!htbp]
\centering
\small
\setlength{\tabcolsep}{3pt}
\caption{Additional strong-LLM requests and tokens in the evaluation pipeline. GCPM G3/G4 rows report the merging stage; standalone branch totals
also include shared proxy generation. Proxy input-token counts cover text only.}
\label{tab:external-llm-cost}
\begin{tabular}{lrrr}
\toprule
Method & Requests & Input tokens & Output tokens \\
\midrule
ABC & 0 & 0 & 0 \\
CLAY & 2,056 & 2,427,707 & 1,118,584 \\
Multi-Sub & 2,042 & 1,922,029 & 856,531 \\
ITGC & 6,265 & 12,273,548 & 1,969,358 \\
GCPM G1/2 & 14,998 & 7,179,554 & 1,421,060 \\
GCPM G3 & 12,695 & 6,128,699 & 1,006,614 \\
GCPM G4 & 12,675 & 6,180,730 & 1,022,508 \\
\bottomrule
\end{tabular}
\par\smallskip
{\footnotesize
G1/2 is the shared proxy-generation stage; G3/G4 rows count merging only. Including the shared stage yields 27,693/27,673 recorded generation requests for standalone G3/G4 branches. Proxy tokens are reconstructed from archived text and exclude image tokens; merge tokens are recorded at runtime.
}
\end{table}

\paragraph{Recovered geometry and validation.}
We matched frozen input/label hashes to the run manifests and reproduced all saved per-unit V-measure, ARI, and coverage scores. For ABC, CLAY, ITGC, and Multi-Sub, replaying fixed-parameter HDBSCAN on the saved vectors reproduced all 449 partitions per method. GCPM vectors and metadata match the original MST source hashes; final predictions are aligned by sorted item ID within each unit. Counts in Table~\ref{tab:embedding-main} exclude noise clusters; noise objects remain in all geometry calculations.

For each task, Alignment is the class-balanced mean squared Euclidean distance between distinct same-class L2-normalized vectors, hence twice $D_{\mathrm{in}}$. We then average over tasks, matching the alignment weighting in the pooled main table. Define $U_{\mathrm{task}}$ as the task-macro mean of $\log\operatorname{mean}_{i\ne j}\exp(-2\|z_i-z_j\|_2^2)$ within each task. $U_{\mathrm{pool}}$ applies the same kernel to all distinct record pairs in the full 14,997-record pool, matching the pooled main-table uniformity definition. Both are exact float64 calculations without pair sampling; they are different aggregation scopes and must not be substituted for each other.

Pooled uniformity is reported for ABC and the GCPM branches in their common encoder coordinates, and descriptively for CLAY's saved ambient coordinates after task-conditioned projection. ITGC's axes refer to task-specific concepts, while Multi-Sub independently optimizes task-local representations; a common cross-task metric space has not been established for these runs. Their pooled uniformity is therefore marked N/A rather than computing arbitrary cross-task distances. This does not prevent within-task geometry or task-macro Alignment. CLAY uses pool-conditioned alignment, so its pooled statistic is not evidence of a globally learned shared geometry. GCPM's semantic merges change labels, not these vectors, and its geometric scores do not measure the effects of the subsequent LLM decisions. No pooled clustering scores are inferred from these statistics.

\begin{table}[!htbp]
\centering
\small
\setlength{\tabcolsep}{3pt}
\caption{External-method geometry recovered from saved representations. All task-wise quantities average the same 449 complete units.}
\label{tab:external-geometry}
\begin{tabular}{lrrrr}
\toprule
Method & $R_{\mathrm{dist}}\downarrow$ & Alignment & $U_{\mathrm{task}}$ & $U_{\mathrm{pool}}$ \\
\midrule
GCPM G1 & 0.7817 & 0.2116 & -0.5581 & -1.9754 \\
GCPM G2 & 0.7797 & 0.3690 & -0.9585 & -2.9360 \\
GCPM G3 & 0.7817 & 0.2116 & -0.5581 & -1.9754 \\
GCPM G4 & 0.7797 & 0.3690 & -0.9585 & -2.9360 \\
ABC & \textbf{0.7184} & 0.0915 & -0.3987 & -3.6362 \\
CLAY & 0.8188 & 1.5644 & -3.1463 & -3.9322 \\
ITGC & 0.8613 & 0.0531 & -0.1564 & N/A \\
Multi-Sub & 1.1604 & 0.1755 & -0.2849 & N/A \\
\bottomrule
\end{tabular}
\end{table}

\paragraph{The inference advantage is architectural, not a measured runtime claim.}
CIM-Embed requires no additional generative strong-LLM calls, as is also true for ABC. The recorded pipelines average 4.58, 4.55, and 13.95 requests per task for CLAY, Multi-Sub, and ITGC, respectively. GCPM's shared proxy-generation stage contains 14,998 archived sample-level generation records, including one supplementary generation. G3 and G4 additionally incur 12,695 and 12,675 recorded merge requests, respectively, including their final recovery runs. Attributing the shared proxy stage to each standalone branch gives 27,693 and 27,673 recorded generation requests for G3 and G4. Proxy token counts are reconstructed from archived text and exclude image tokens, whereas merge token counts are recorded at runtime (Appendix Table~\ref{tab:external-llm-cost}). These counts characterize the generative workload rather than end-to-end latency or memory. The quality comparison remains system-level because input modalities, rejection policies, and clustering-parameter selection differ across methods.

\subsection{Related-Work Context}
\label{app:related-context}

\paragraph{Conditional similarity and retrieval.}
Shared multimodal representations support cross-modal retrieval, with instructions specifying the current matching task or similarity criterion~\citep{uniir2024,mmembed2025,jiang2025vlm2vec,qwen3vlembedding2026}. Conditional subspaces, instruction-tuned text embeddings, and conditional similarity evaluations study this control from complementary perspectives~\citep{veit2017csn,su2023one,promptriever2025,deshpande2023csts,vaze2023genecis}. These approaches ask which objects are relevant or should be similar; our task additionally requires recovering a complete partition of the same object pool under each grouping instruction.

\paragraph{Conditional representation adaptation and transformation.}
A second family adjusts representations so that matching or clustering reflects a specified condition. ABC adapts the model and CLAY applies text-conditioned projection for visual matching; InBedder learns instruction-following text representations through question-answering supervision, whereas GSTransform transforms existing representations~\citep{schneider2025abc,lim2026clay,peng2024inbedder,feng2025gstransform}. Multi-MaP and Multi-Sub learn proxies corresponding to different semantic perspectives~\citep{yao2024multimap,yao2024multisub}. CIM-Embed likewise applies control at the representation level, but uses instruction-defined co-membership of image--text objects as multi-positive supervision and studies structured sampling and within-task versus cross-task comparisons.

\paragraph{Generative and reasoning-based clustering.}
Conditions can also enter clustering through text rewriting, relational feedback, or generated assignments~\citep{pangnaylor2025controllable,zhang2023clusterllm,qing2026clusterr1}. In visual clustering, IC$|$TC and ITGC use generated descriptions or concepts to guide grouping; GCPM encodes guideline-aware descriptions, with its accompanying framework using language models to merge clusters~\citep{kwon2024ictc,zhao2025itgc,zhong2026guidelineclustering}. Relative to these three routes, our focus is complete partition recovery rather than relevance ranking alone, direct adaptation of object encoders through group supervision, and inference using conditional embeddings and a standard clustering algorithm without intermediate generation or language-model cluster refinement.

\paragraph{Contrastive supervision and batch sampling.}
InfoNCE contrasts an anchor's positive against negatives, while supervised contrastive learning supports multiple same-class positives~\citep{oord2018cpc,khosla2020supervised}. Batch sampling determines which examples are included in each training step and thus which positive and negative pairs can be formed for contrastive learning. Class-balanced instance sampling supports within-batch triplet mining~\citep{hermans2017defense}, and Multi-Similarity Loss studies pair selection and weighting through gradient analysis~\citep{wang2019multisimilarity}. Building on these contrastive learning objectives and sampling strategies, CIM-Embed defines positive pairs by group membership under a given instruction and constructs batches spanning multiple grouping tasks. We further examine whether negatives are drawn only from the same task or also from other tasks.
\FloatBarrier

\section{Cross-Domain Transfer}
\label{app:additional-transfer}
\subsection{Cards}

\paragraph{Data and conditions.} Cards contains 8,029 images evaluated under two grouping conditions: rank, with 13 reference classes, and suit, with four. Table~\ref{tab:cards-grid12} compares 32 model configurations on this shared image pool. Each configuration produces full-native-dimensional embeddings under rank instructions, suit instructions, and an image-only condition, yielding 24,087 vectors. Labels, card names, and class-bearing paths are excluded from the model text.

\paragraph{Frozen models and selection.} All model weights are frozen, without training on Cards. Clustering uses L2-normalized embeddings and Euclidean HDBSCAN (\texttt{hdbscan} 0.8.44, EOM), without UMAP. The 12-point grid combines $\texttt{min\_cluster\_size}\in\{2,3,4,5\}$ and $\texttt{min\_samples}\in\{\mathrm{None},1,2\}$; \texttt{None} sets the minimum samples to the minimum cluster size. One shared parameter pair is selected per model by the equal-weight mean rank/suit V-measure, with ties resolved by higher mean coverage and then fixed grid order. The image-only condition reuses the selected pair, and its results are included in the archived artifacts. Cards labels are used for both parameter selection and scoring; the results therefore characterize frozen-model transfer under target-label-assisted clustering selection. Noise label $-1$ is retained as one predicted cluster, and coverage is the fraction of objects assigned to non-noise clusters. Mean V-measure, ARI, and coverage equally weight the two instructed conditions.

\paragraph{Results.} Qwen 8B Direct Within achieves the highest mean V-measure among the evaluated configurations (88.55), exceeding the strongest general-purpose baseline, Qwen3-VL 2B (56.36), by 32.19 percentage points. Its rank and suit scores are 88.80 and 88.29, respectively, showing strong grouping quality under both conditions. WeMM 2B Direct Within follows with a mean V-measure of 87.75, while WeMM 9B Direct Within achieves the highest rank V-measure of 92.90 (Table~\ref{tab:cards-grid12}). These results extend the observed benefits of product-trained conditional representations to card grouping. Performance varies across training configurations: Qwen 8B Within + replay achieves a rank V-measure of 77.95 but a suit V-measure of 0.53.

\begin{table}[!htbp]
\centering
\caption{Cards conditional clustering with a 12-point grid selected using target labels.}
\label{tab:cards-grid12}
\begingroup
\scriptsize
\setlength{\tabcolsep}{2pt}
\resizebox{\linewidth}{!}{%
\begin{tabular}{@{}lrrrrrl@{}}
\toprule
Model & Rank $V$ & Suit $V$ & Mean $V$ & ARI & Cov. & $(c,s)$ \\
\midrule
Qwen 8B Direct Within & 88.80 & \textbf{88.29} & \textbf{88.55} & \textbf{0.8748} & \textbf{94.53} & (5, --) \\
WeMM 2B Direct Within & 88.55 & 86.94 & 87.75 & 0.8703 & 92.13 & (5, --) \\
Earlier MRL Cross & 88.18 & 86.00 & 87.09 & 0.8551 & 93.29 & (5, --) \\
Qwen 8B Direct Cross & 87.10 & 85.00 & 86.05 & 0.8359 & 91.93 & (5, --) \\
WeMM 9B Cross + replay & 85.32 & 83.74 & 84.53 & 0.8042 & 88.50 & (5, --) \\
WeMM 9B Direct Within & \textbf{92.90} & 75.44 & 84.17 & 0.8320 & 90.32 & (5, --) \\
Qwen 2B Direct Cross & 86.33 & 80.21 & 83.27 & 0.8029 & 88.49 & (5, 2) \\
Qwen 2B Direct InfoNCE & 85.76 & 78.86 & 82.31 & 0.7356 & 91.36 & (5, --) \\
WeMM 2B Cross + replay & 79.46 & 84.37 & 81.91 & 0.7427 & 86.77 & (4, --) \\
Qwen 2B Direct Within & 84.79 & 78.25 & 81.52 & 0.7605 & 86.52 & (5, --) \\
WeMM 2B Direct Cross & 88.82 & 73.62 & 81.22 & 0.8024 & 86.39 & (5, --) \\
Qwen 2B Cross + replay & 88.60 & 68.37 & 78.49 & 0.7121 & 82.92 & (5, --) \\
Qwen 8B InfoNCE Stage3 + replay & 88.39 & 67.34 & 77.87 & 0.7238 & 84.62 & (4, --) \\
WeMM 9B Direct Cross & 91.14 & 60.85 & 75.99 & 0.7087 & 87.36 & (5, --) \\
Qwen 8B Direct InfoNCE & 88.81 & 62.10 & 75.45 & 0.7012 & 82.55 & (5, --) \\
Qwen 8B Cross + replay & 74.33 & 73.72 & 74.02 & 0.5832 & 75.76 & (5, --) \\
WeMM 9B Direct InfoNCE & 84.22 & 60.36 & 72.29 & 0.6440 & 81.01 & (5, --) \\
WeMM 2B Direct InfoNCE & 88.76 & 37.06 & 62.91 & 0.4315 & 73.51 & (5, --) \\
Qwen3-VL 2B (official) & 74.33 & 38.39 & 56.36 & 0.2982 & 66.07 & (5, --) \\
WeMM 9B (official) & 49.73 & 33.14 & 41.43 & 0.0554 & 53.77 & (5, 1) \\
Qwen3-VL 8B (official) & 48.90 & 33.56 & 41.23 & 0.0433 & 51.30 & (4, 2) \\
Qwen 8B Within + replay & 77.95 & 0.53 & 39.24 & 0.3134 & 91.11 & (5, 1) \\
WeMM 2B (official) & 48.48 & 28.07 & 38.28 & 0.0503 & 50.81 & (4, 1) \\
Tongyi Plus & 46.58 & 13.54 & 30.06 & 0.0409 & 52.57 & (5, 1) \\
Tongyi Flash & 36.55 & 22.22 & 29.39 & 0.0033 & 54.09 & (2, 1) \\
WeMM 9B Hier & 35.07 & 23.48 & 29.28 & 0.0018 & 51.91 & (2, 1) \\
Qwen 8B InfoNCE Stage2 & 33.73 & 22.62 & 28.18 & 0.0020 & 51.96 & (2, 1) \\
Gemini Embedding 2 & 34.64 & 20.25 & 27.44 & 0.0029 & 53.59 & (2, 1) \\
Qwen 8B Hier & 33.31 & 21.34 & 27.33 & 0.0017 & 52.22 & (2, 1) \\
WeMM 2B Hier & 30.99 & 20.13 & 25.56 & 0.0008 & 46.34 & (2, 1) \\
Qwen 2B Hier & 29.91 & 20.66 & 25.28 & 0.0006 & 48.73 & (2, 1) \\
Qwen 8B InfoNCE Stage1 & 29.41 & 17.68 & 23.55 & 0.0009 & 48.73 & (2, 1) \\
\bottomrule
\end{tabular}}
\endgroup
\par\smallskip
\begin{minipage}{\linewidth}
\scriptsize
V and coverage are percentages; ARI is unscaled. $(c,s)$ denotes minimum cluster size and minimum samples; -- means None. Flash/Plus use corrected fused image--text inputs, replacing invalid text-only outputs. The MRL model is a separate checkpoint configuration; the other trained entries use their final checkpoints.
\end{minipage}
\end{table}

\subsection{CLEVR-4 and CIFAR10-MC}
\paragraph{Data and conditions.}
CLEVR-4 uses all 10,531 images from the 10k release, combining the original 8,424 training and 2,107 validation images without training the encoder on them. Color, shape, texture, and count each define ten reference groups and are evaluated separately. CIFAR10-MC uses all 50,000 available $32\times32$ images in its release, aligning both label organizations to the same pool by image hashes. Type separates animals from transportation; Environment defines Land, Air, and Water by the main object's typical association, with frogs assigned to Water by the released labels. The package does not provide a directly usable original train/test split. Native model processors resize the original images without super-resolution enhancement.

\paragraph{Frozen models and selection.}
Both datasets use the same ten frozen checkpoints, without target-data training or checkpoint selection. Configuration names describe the training objective; identical names across datasets do not establish checkpoint identity with the Cards evaluation. Unless a backbone is specified, trained models use Qwen3-VL-Embedding 8B. Qwen 2B and WeMM 2B use 2,048 dimensions; the others use 4,096. HDBSCAN uses Euclidean distances on normalized full-dimensional vectors without reference cluster counts. The grid combines $\texttt{min\_samples}\in\{1,2,5,10,20,50\}$ with $\texttt{min\_cluster\_size}\in\{2,5,10,20,50,100,200,500\}$. Each model selects one pair per dataset by mean conditional V, breaking ties by mean coverage and then grid order; all conditions share that pair. Full-pool labels serve both selection and scoring, introducing optimistic selection bias rather than an independently selected test estimate. All objects remain in scoring, with noise $-1$ treated as one predicted cluster; non-noise coverage and cluster counts are reported separately. Dataset means equally weight four or two conditions.

\paragraph{Mixed training and metrics.}
In the CLEVR-4 and CIFAR10-MC tables, ``Mix'' denotes generic-data replay: 18,000 instruction batches of 20 object positions are mixed with 6,000 attribute-agnostic product batches of 16 positions. The 3:1 ratio refers to batch counts. All encoders are evaluated without training on the target datasets. ``Stage2'' denotes the attribute-agnostic training checkpoint, which receives grouping instructions during conditional evaluation. 
$Q$ is full-pool cosine-density-tree $Q_{\mathrm{target}}$ with $k=5,T=1$. For $R_{\mathrm{dist}}=D_{\mathrm{in}}/D_{\mathrm{out}}$, the denominator averages each reference class's mean cross-class distance to its nearest other class, not centroid distances. Alignment equals $2D_{\mathrm{in}}$; Uniformity uses the log mean exponentiated negative twice squared distance over all distinct normalized-vector pairs. Metrics are computed per condition and then averaged; geometry metrics are not used for selection.

\paragraph{Instruction controls.}
Controls hold reference labels, images, and selected parameters fixed. CLEVR-4's instruction-free column averages scores against all four label sets; its other-instruction difference subtracts the mean V under the three alternative instructions from correct-instruction V, then averages over targets. CIFAR10-MC subtracts V under the other valid instruction and has no instruction-free input evaluation. Differences retain the source's unrounded calculations before conversion to percentage points, so they may differ slightly from subtraction of displayed rounded scores.

\begin{table}[htbp]
\centering
\caption{CLEVR-4: equal-weight means over four conditions. V and its differences are in percentage points; coverage is a percentage.}
\label{tab:clevr4-transfer}
\begingroup\scriptsize
\setlength{\tabcolsep}{2pt}
\begin{tabular*}{\linewidth}{@{\extracolsep{\fill}}l*{8}{r}@{}}
\toprule
Model & $V$ & ARI & $Q$ & Cov. & Clusters & $R_{\mathrm{dist}}$ & Align. & Unif. \\
\midrule
Qwen 8B Base & 42.07 & 0.1701 & 0.0308 & 84.94 & 517.50 & 0.8218 & 0.5564 & -1.4372 \\
Direct InfoNCE & 60.93 & 0.3298 & 0.1991 & 95.45 & 28.00 & 0.5015 & 0.2233 & -1.1758 \\
Direct Within & 64.19 & 0.4796 & 0.2094 & 84.15 & 10.00 & 0.3822 & 0.0613 & -0.6255 \\
Direct Cross & 54.09 & 0.2826 & 0.1705 & 93.16 & 42.00 & 0.4934 & 0.1721 & -0.9979 \\
Stage2 InfoNCE & 35.41 & 0.1531 & 0.1132 & 84.34 & 1006.75 & 0.8008 & 0.5547 & -1.3047 \\
Within + Mix & 66.46 & 0.4702 & 0.2038 & 91.25 & 14.50 & 0.3676 & 0.0638 & -0.6026 \\
Cross + Mix & 58.94 & 0.3010 & 0.1586 & 90.36 & 45.50 & 0.4562 & 0.1270 & -0.8226 \\
Qwen 2B Cross + Mix & 60.06 & 0.3164 & 0.1574 & 90.12 & 70.75 & 0.4507 & 0.1174 & -0.8188 \\
WeMM 2B Cross + Mix & 48.59 & 0.2500 & 0.1631 & 88.32 & 961.25 & 0.5510 & 0.1908 & -0.9425 \\
WeMM 9B Cross + Mix & 52.14 & 0.2866 & 0.1478 & 85.45 & 1064.50 & 0.5321 & 0.1787 & -0.9066 \\
\bottomrule
\end{tabular*}
\endgroup
\end{table}

\begin{table}[htbp]
\centering
\caption{CLEVR-4: instruction controls on the fixed image pool. V and its differences are in percentage points; coverage is a percentage.}
\label{tab:clevr4-controls}
\begingroup\scriptsize
\setlength{\tabcolsep}{2pt}
\begin{tabular*}{\linewidth}{@{\extracolsep{\fill}}l*{4}{r}@{}}
\toprule
Model & $V_{\mathrm{correct}}$ & $V_{\mathrm{none}}$ & $\Delta V_{\mathrm{none}}$ & $\Delta V_{\mathrm{other}}$ \\
\midrule
Qwen 8B Base & 42.07 & 35.86 & +6.21 & +11.81 \\
Direct InfoNCE & 60.93 & 23.83 & +37.10 & +48.05 \\
Direct Within & 64.19 & 7.61 & +56.58 & +59.22 \\
Direct Cross & 54.09 & 24.78 & +29.31 & +35.42 \\
Stage2 InfoNCE & 35.41 & 31.33 & +4.08 & +5.22 \\
Within + Mix & 66.46 & 23.94 & +42.53 & +60.18 \\
Cross + Mix & 58.94 & 23.97 & +34.97 & +41.80 \\
Qwen 2B Cross + Mix & 60.06 & 24.40 & +35.67 & +42.01 \\
WeMM 2B Cross + Mix & 48.59 & 29.96 & +18.64 & +23.50 \\
WeMM 9B Cross + Mix & 52.14 & 31.63 & +20.50 & +27.22 \\
\bottomrule
\end{tabular*}
\endgroup
\end{table}

\begin{table}[htbp]
\centering
\caption{CIFAR10-MC: equal-weight means over two conditions. V and its differences are in percentage points; coverage is a percentage.}
\label{tab:cifar10mc-transfer}
\begingroup\scriptsize
\setlength{\tabcolsep}{2pt}
\begin{tabular*}{\linewidth}{@{\extracolsep{\fill}}l*{8}{r}@{}}
\toprule
Model & $V$ & ARI & $Q$ & Cov. & Clusters & $R_{\mathrm{dist}}$ & Align. & Unif. \\
\midrule
Qwen 8B Base & 12.44 & 0.0042 & 0.0235 & 38.55 & 7246.50 & 0.8750 & 1.1154 & -2.3130 \\
Direct InfoNCE & 32.33 & 0.1413 & 0.0555 & 44.46 & 9.00 & 0.7445 & 0.5569 & -1.2517 \\
Direct Within & 31.76 & 0.1227 & 0.0751 & 48.84 & 8.50 & 0.6287 & 0.2241 & -0.5551 \\
Direct Cross & 31.75 & 0.1033 & 0.0451 & 49.03 & 8.50 & 0.6375 & 0.4023 & -0.9462 \\
Stage2 InfoNCE & 13.35 & 0.0159 & 0.0286 & 21.76 & 3.00 & 0.7732 & 0.3639 & -0.7788 \\
Within + Mix & 27.28 & 0.0860 & 0.0631 & 42.31 & 10.50 & 0.6357 & 0.2073 & -0.5037 \\
Cross + Mix & 28.02 & 0.0964 & 0.0853 & 38.63 & 10.50 & 0.6484 & 0.2697 & -0.6360 \\
Qwen 2B Cross + Mix & 22.71 & 0.1325 & 0.0521 & 36.05 & 5.50 & 0.6499 & 0.2704 & -0.6457 \\
WeMM 2B Cross + Mix & 23.88 & 0.1481 & 0.0283 & 34.02 & 4.00 & 0.6652 & 0.3248 & -0.7497 \\
WeMM 9B Cross + Mix & 34.16 & 0.1516 & 0.0343 & 50.23 & 6.50 & 0.6489 & 0.3267 & -0.7535 \\
\bottomrule
\end{tabular*}
\endgroup
\end{table}

\begin{table}[htbp]
\centering
\caption{CIFAR10-MC: per-condition results and cross-instruction controls. V and its differences are in percentage points; coverage is a percentage.}
\label{tab:cifar10mc-controls}
\begingroup\scriptsize
\setlength{\tabcolsep}{2pt}
\begin{tabular*}{\linewidth}{@{\extracolsep{\fill}}l*{4}{r}@{}}
\toprule
Model & $V_{\mathrm{Type}}$ & $V_{\mathrm{Env.}}$ & $\Delta V_{\mathrm{Type}}$ & $\Delta V_{\mathrm{Env.}}$ \\
\midrule
Qwen 8B Base & 10.86 & 14.02 & -0.03 & +0.04 \\
Direct InfoNCE & 33.01 & 31.65 & +0.60 & -1.53 \\
Direct Within & 29.74 & 33.79 & -6.51 & +0.73 \\
Direct Cross & 29.79 & 33.72 & -6.37 & +0.27 \\
Stage2 InfoNCE & 20.76 & 5.93 & +10.99 & -4.89 \\
Within + Mix & 26.26 & 28.29 & -4.66 & -0.69 \\
Cross + Mix & 27.21 & 28.84 & -3.58 & +0.63 \\
Qwen 2B Cross + Mix & 38.29 & 7.14 & +27.86 & -10.97 \\
WeMM 2B Cross + Mix & 36.64 & 11.11 & +25.69 & -7.47 \\
WeMM 9B Cross + Mix & 36.33 & 31.99 & +7.28 & +0.83 \\
\bottomrule
\end{tabular*}
\endgroup
\end{table}
\clearpage

\clearpage
\subsection{Instruction Response and Transfer Analysis}
\label{app:transfer-diagnostics}

\paragraph{Conditional grouping transfers beyond the training domain.}
Product-trained encoders can recover meaningful non-product partitions without updating their weights on Cards. Across 32 configurations evaluated on the same 8,029 images, Qwen 8B Direct Within reaches mean V=88.55 and ARI=0.8748, compared with 56.36 and 0.2982 for Qwen3-VL 2B, the strongest general-purpose baseline in this comparison. WeMM 2B Direct Within also reaches V=87.75 and ARI=0.8703 (Appendix Table~\ref{tab:cards-grid12}). The effect is therefore observed in more than one backbone.

\paragraph{Successful transfer covers both instructions, but is configuration-dependent.}
Qwen 8B Direct Within obtains rank/suit V=88.80/88.29, and WeMM 2B Direct Within obtains 88.55/86.94. Their strong averages reflect recovery of both partitions on the shared image pool. However, Qwen 8B Within with replay reaches rank V=77.95 but suit V=0.53, placing 7,828 of 8,029 images in one mixed suit cluster. Product-category retention therefore does not guarantee successful cross-domain conditional grouping. For Cards, all models use the same 12-point HDBSCAN search protocol, with one parameter pair selected per model for both tasks using Cards labels. These results establish cross-domain representation utility under target-label-selected clustering, rather than frozen-parameter zero-shot clustering.

\paragraph{CLEVR-4 supports transfer of visible-attribute and count conditions.}
CLEVR-4 groups scenes by color, shape, surface texture, or object count. Among ten frozen checkpoints, Within with mixed training obtains the highest mean V at 66.46, versus 42.07 for Qwen 8B Base; Direct Within has the highest ARI at 0.4796 (Appendix Table~\ref{tab:clevr4-transfer}). Within with mixed training reaches color/shape V=90.07/95.08, but texture/count V=41.83/38.88. Holding images and selected parameters fixed, its macro-average gains over no instruction and other valid instructions are 42.53 and 60.18 percentage points (Appendix Table~\ref{tab:clevr4-controls}). These controls provide evidence of conditional response beyond a before/after training comparison, although parameters are still selected using full-pool labels under the correct conditions.

\paragraph{CIFAR10-MC separates representation gains from conditional response.}
CIFAR10-MC groups the same images by broad object type or the object's typical associated environment; the latter follows the released semantic labels rather than the photographed background. On the same Qwen 8B backbone, Direct InfoNCE raises mean V from the Base model's 12.44 to 32.33. WeMM 9B Cross with mixed training has the highest evaluated mean V at 34.16, but ARI is only 0.1516 and coverage is 50.23\% (Appendix Table~\ref{tab:cifar10mc-transfer}). Most configurations do not benefit from the correct instruction on both conditions. WeMM 9B's Type and Environment gains over the alternative instruction are 7.28 and 0.83 percentage points, without repeated-run confidence intervals (Appendix Table~\ref{tab:cifar10mc-controls}). No instruction-free input control was extracted. Both Qwen 8B mixed variants also underperform their corresponding unmixed variants, reversing the CLEVR-4 trend. Better representations therefore do not automatically imply robust conditional switching, and the transfer benefit of mixed training depends on the dataset.

Together, the three datasets show that frozen product-trained encoders support grouping under some out-of-domain conditions, with outcomes depending on dataset, attribute, and training configuration. Parameters are selected separately using each target dataset's labels; results are neither pooled into a cross-dataset overall score nor treated as frozen-parameter zero-shot tests.
\FloatBarrier

\section{Qualitative Visualizations}
\label{app:representation-visualization}
\begin{figure}[!htbp]
\centering
\includegraphics[width=\linewidth,height=0.48\textheight,keepaspectratio]{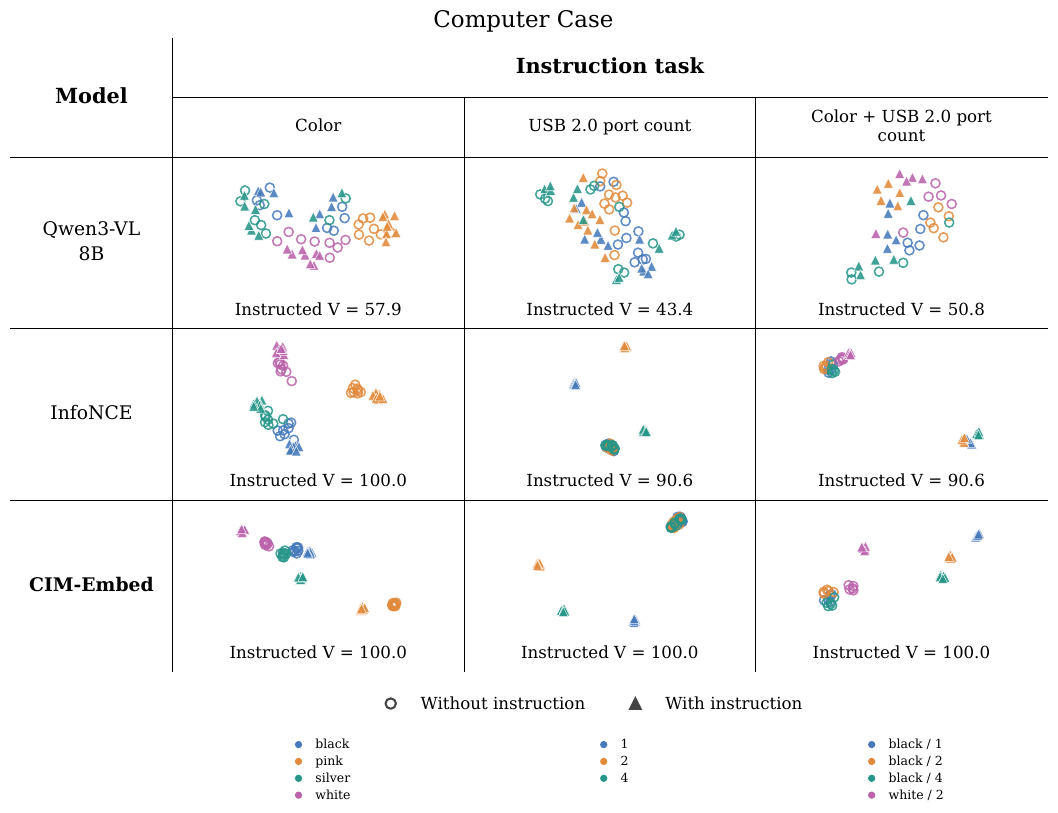}
\caption{\textbf{Computer Case representations with and without instructions.} Models, projections, and evaluation conventions follow Figure~\ref{fig:instruction-umap-wheel}. Hollow circles and filled triangles denote representations without and with instructions, respectively; colors denote ground-truth groups. Both conditions are jointly projected within each cell, with independent fits across cells. Product pools are identical across models within a task but can differ across tasks. Instructed $V$ is original-space HDBSCAN V-measure under the evaluation-label-tuned protocol.}
\label{fig:instruction-umap-computer-case}
\end{figure}

\begin{figure}[!htbp]
\centering
\includegraphics[width=\linewidth,height=0.48\textheight,keepaspectratio]{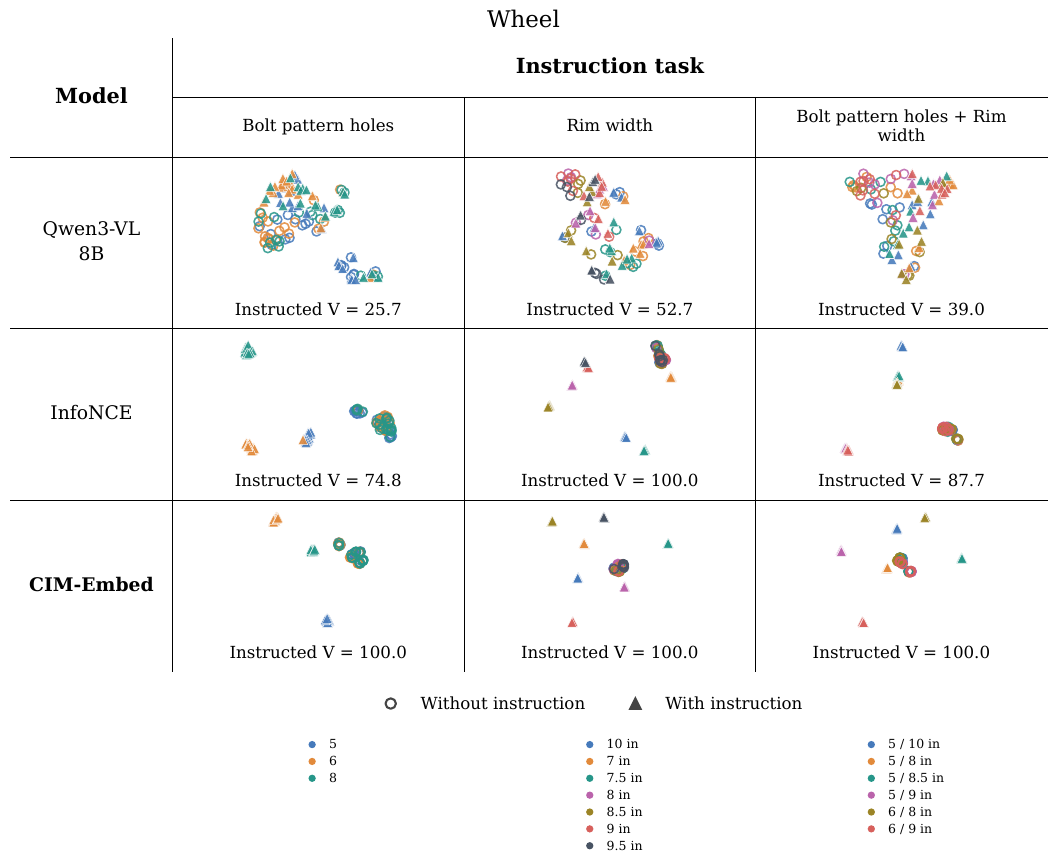}
\caption{\textbf{Real representations with and without instructions.} Single- and joint-attribute tasks on Wheel. Rows denote models and columns denote single- or joint-attribute tasks. Hollow circles indicate no instruction, filled triangles indicate an instruction, and colors denote ground-truth groups. Within each cell, the two conditions are jointly projected with UMAP after L2 normalization; projections are fitted independently across models and tasks, so positions and distances should not be compared across cells. Each task uses the same product pool across models, but pools can differ across tasks. Instructed $V$ is HDBSCAN V-measure (0--100) on the original instructed 4,096-dimensional representations under the evaluation-label-tuned protocol used in the main tables, not a score computed from the two-dimensional points.}
\label{fig:instruction-umap-wheel}
\end{figure}

Figure~\ref{fig:instruction-umap-computer-case} supplements Figure~\ref{fig:instruction-umap-wheel} with the color, USB 2.0 port count, and joint-attribute tasks on Computer Case. This example is also selected for strong SupCon scores and serves as a qualitative illustration.

\paragraph{Visual examples illustrate the recovered group structure.}
On Wheel, instructed Cross embeddings form compact target groups and achieve original-space V=100 on all three illustrated tasks, while InfoNCE retains partition errors on some tasks (Figure~\ref{fig:instruction-umap-wheel}; Computer Case appears in Appendix Figure~\ref{fig:instruction-umap-computer-case}). This illustrates what improved partition recovery can look like, rather than establishing its frequency: the examples were selected for strong SupCon scores, and independently fitted two-dimensional panels cannot be compared by absolute position.

\FloatBarrier

\section{Extended Limitations}
\label{app:extended-limitations}
\subsection{Data Coverage and Annotation}
Model-assisted labels can retain errors despite metadata checks. Category-held-out evaluation tests 50 leaf categories within product data. Balanced pools of 10--98 products reduce natural frequency variation and do not establish performance on large, long-tailed catalogs.

\subsection{Evaluation and Input Controls}
HDBSCAN selection uses evaluation labels, including target-domain labels in transfer, so the results are not independently selected held-out estimates. Product no-instruction comparisons share selected clustering parameters, but the available records do not establish identical templates and missing-data handling for every checkpoint. The WeMM 9B input audit does not resolve these differences for other models. Exact relation switching on matched product pools is not established. Pooled V-measure and ARI quantify task-specific partition recovery after joint clustering; assessing cross-task cluster identity requires an additional global evaluation.
Repeated-seed uncertainty and a matched complete-assignment comparison are not reported.

\subsection{Training Attribution}
The SupCon comparisons change both loss and sampling, and staged configurations change both initialization history and replay. These results characterize complete configurations rather than isolated component effects. Complete run-level training manifests and template mappings are unavailable for some compared checkpoints; the released corpus counts do not specify every model's training exposure. Category retention does not establish retention of retrieval ability.

\subsection{Generalization}
The three transfer datasets support some out-of-domain grouping, but do not establish universal instruction switching: CLEVR-4 shows stronger conditional response than CIFAR10-MC. New attribute combinations, natural long-tailed pools, and cross-type grouping are outside the evaluated scope. Cross-type tasks would require different pools, labels, and contrastive masks. Target-label parameter selection further limits conclusions about deployment without target labels.
\FloatBarrier
\clearpage

\end{document}